\documentclass{article}

\usepackage{microtype}
\usepackage{graphicx}
\usepackage{subfigure}
\usepackage{booktabs} 

\usepackage{hyperref}

\usepackage[accepted]{icml2025}

\usepackage{amsmath}
\usepackage{amssymb}
\usepackage{mathtools}
\usepackage{amsthm}

\usepackage{booktabs}       
\usepackage{amsfonts}       
\usepackage{nicefrac}       
\usepackage{microtype}      
\usepackage{xcolor}         
\usepackage{siunitx}
\usepackage{cuted}
\usepackage{capt-of}
\usepackage{tcolorbox}
\usepackage{amsmath}
\usepackage{amssymb}
\usepackage{multirow}
\usepackage{subcaption}
\usepackage{xspace}
\usepackage{wrapfig,lipsum}      
\usepackage{colortbl}
\usepackage{array}

\definecolor{hookersgreen}{rgb}{0.0, 0.44, 0.0}
\definecolor{indiagreen}{rgb}{0.07, 0.53, 0.03}
\definecolor{islamicgreen}{rgb}{0.0, 0.56, 0.0}
\definecolor{kellygreen}{rgb}{0.3, 0.73, 0.09}
\definecolor{alizarin}{rgb}{0.82, 0.1, 0.26}

\usepackage{pifont}
\newcommand{\cmark}{{\color{kellygreen} \ding{51}}}
\newcommand{\xmark}{{\color{alizarin} \ding{55}}}

\definecolor{Gray}{gray}{0.90}
\definecolor{lightgrey}{gray}{0.965}
\definecolor{middlegrey}{gray}{0.95}
\definecolor{lightcyan}{RGB}{178, 255, 255}
\definecolor{lightpurple}{RGB}{203, 195, 227}
\definecolor{lightgreen}{RGB}{224,255,210}
\definecolor{lightyellow}{RGB}{255,255,204}

\definecolor{rank1}{rgb}{0.87, 0.36, 0.14}  
\definecolor{rank2}{rgb}{0.90, 0.45, 0.18}
\definecolor{rank3}{rgb}{0.92, 0.55, 0.22}
\definecolor{rank4}{rgb}{0.94, 0.65, 0.26}
\definecolor{rank5}{rgb}{0.85, 0.65, 0.20}  
\definecolor{rank6}{rgb}{0.80, 0.70, 0.25}  
\definecolor{rank7}{rgb}{0.75, 0.75, 0.30}  
\definecolor{rank8}{rgb}{0.70, 0.80, 0.35}  
\definecolor{rank9}{rgb}{0.70, 0.85, 0.70}  
\definecolor{rank10}{rgb}{0.60, 0.83, 0.75}
\definecolor{rank11}{rgb}{0.50, 0.80, 0.80}
\definecolor{rank12}{rgb}{0.45, 0.78, 0.85}
\definecolor{rank13}{rgb}{0.40, 0.75, 0.90}
\definecolor{rank14}{rgb}{0.35, 0.65, 0.92}
\definecolor{rank15}{rgb}{0.35, 0.55, 0.95}
\definecolor{rank16}{rgb}{0.45, 0.55, 0.95}
\definecolor{rank17}{rgb}{0.55, 0.50, 0.92}
\definecolor{rank18}{rgb}{0.65, 0.50, 0.90}
\definecolor{rank19}{rgb}{0.75, 0.50, 0.85}
\definecolor{rank20}{rgb}{0.85, 0.50, 0.80}  
\definecolor{rank21}{rgb}{0.90, 0.45, 0.75}

\newcommand{\dataset}{\textsc{MMSci}\xspace}

\usepackage[capitalize,noabbrev]{cleveref}

\theoremstyle{plain}

\theoremstyle{definition}

\theoremstyle{remark}

\usepackage[textsize=tiny]{todonotes}

\icmltitlerunning{\dataset: A Dataset for Graduate-Level Multi-Discipline Multimodal Scientific Understanding}

\begin{document}

\twocolumn[
\icmltitle{\dataset: A Dataset for Graduate-Level Multi-Discipline Multimodal \\ Scientific Understanding}



\icmlsetsymbol{equal}{*}

\begin{icmlauthorlist}
\icmlauthor{Zekun Li}{yyy}
\icmlauthor{Xianjun Yang}{yyy}
\icmlauthor{Kyuri Choi}{comp}
\icmlauthor{Wanrong Zhu}{yyy}
\icmlauthor{Ryan Hsieh}{yyy}
\icmlauthor{HyeonJung Kim}{comp}
\icmlauthor{Jin Hyuk Lim}{comp}
\icmlauthor{Sungyoung Ji}{comp}
\icmlauthor{Byungju Lee}{comp,zzz}
\icmlauthor{Xifeng Yan}{yyy}
\icmlauthor{Linda Ruth Petzold}{yyy}
\icmlauthor{Stephen D. Wilson}{yyy}
\icmlauthor{Woosang Lim}{equal,comp}
\icmlauthor{William Yang Wang}{equal,yyy}
\end{icmlauthorlist}


\icmlaffiliation{yyy}{University of California, Santa Barbara}
\icmlaffiliation{zzz}{KIST}
\icmlaffiliation{comp}{POSCO HOLDINGS}

\icmlcorrespondingauthor{Woosang Lim}{woosang\_lim@posco-inc.com}
\icmlcorrespondingauthor{William Yang Wang}{william@cs.ucsb.edu}

\icmlkeywords{Machine Learning, ICML}

\vskip 0.3in
]



\printAffiliationsAndNotice{\icmlEqualContribution} 



\begin{abstract}
Scientific figure interpretation is a crucial capability for AI-driven scientific assistants built on advanced Large Vision Language Models. However, current datasets and benchmarks primarily focus on simple charts or other relatively straightforward figures from limited science domains. To address this gap, we present a comprehensive dataset compiled from peer-reviewed Nature Communications articles covering 72 scientific fields, encompassing complex visualizations such as schematic diagrams, microscopic images, and experimental data which require graduate-level expertise to interpret.
We evaluated 19 proprietary and open-source models on two benchmark tasks, figure captioning and multiple-choice, and conducted human expert annotation. Our analysis revealed significant task challenges and performance gaps among models. Beyond serving as a benchmark, this dataset serves as a valuable resource for large-scale training. Fine-tuning Qwen2-VL-7B with our task-specific data achieved better performance than GPT-4o and even human experts in multiple-choice evaluations. Furthermore, continuous pre-training on our interleaved article and figure data substantially enhanced the model's downstream task performance in materials science. 
We will release our dataset to support further research.

\end{abstract}

\section{Introduction}

\begin{figure}[htbp]
\footnotesize
\centering
\includegraphics[width=\linewidth]{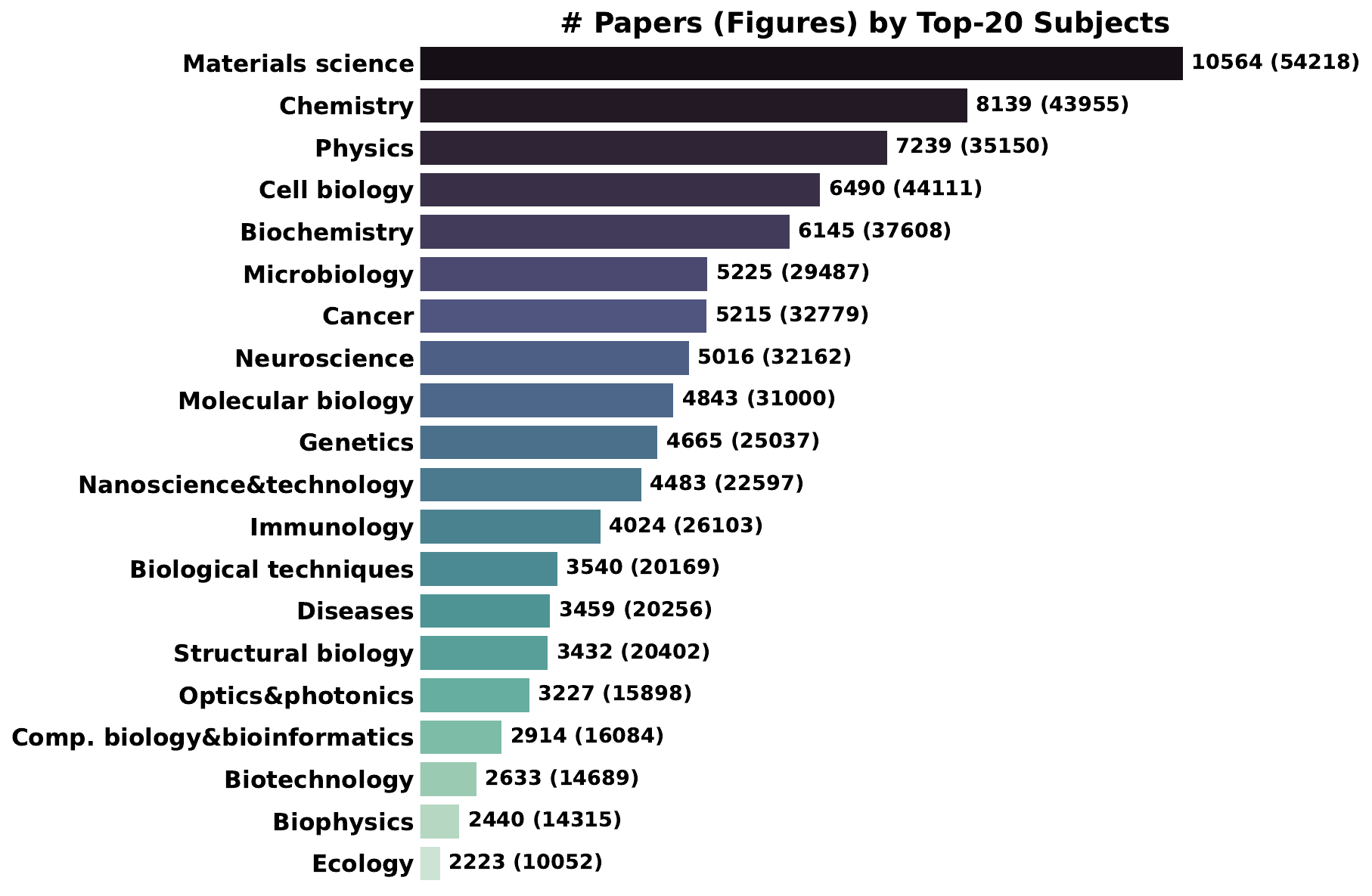}
\caption{\textbf{The top 20 out of \textbf{72 science subjects} with the most articles in our dataset \dataset.} The corresponding numbers of papers and figures (in brackets) are shown.
}\
\vspace{-5mm}
\label{fig:top20}
\end{figure}

\begin{figure*}[htbp]
\footnotesize
\centering
\includegraphics[width=0.98\textwidth]{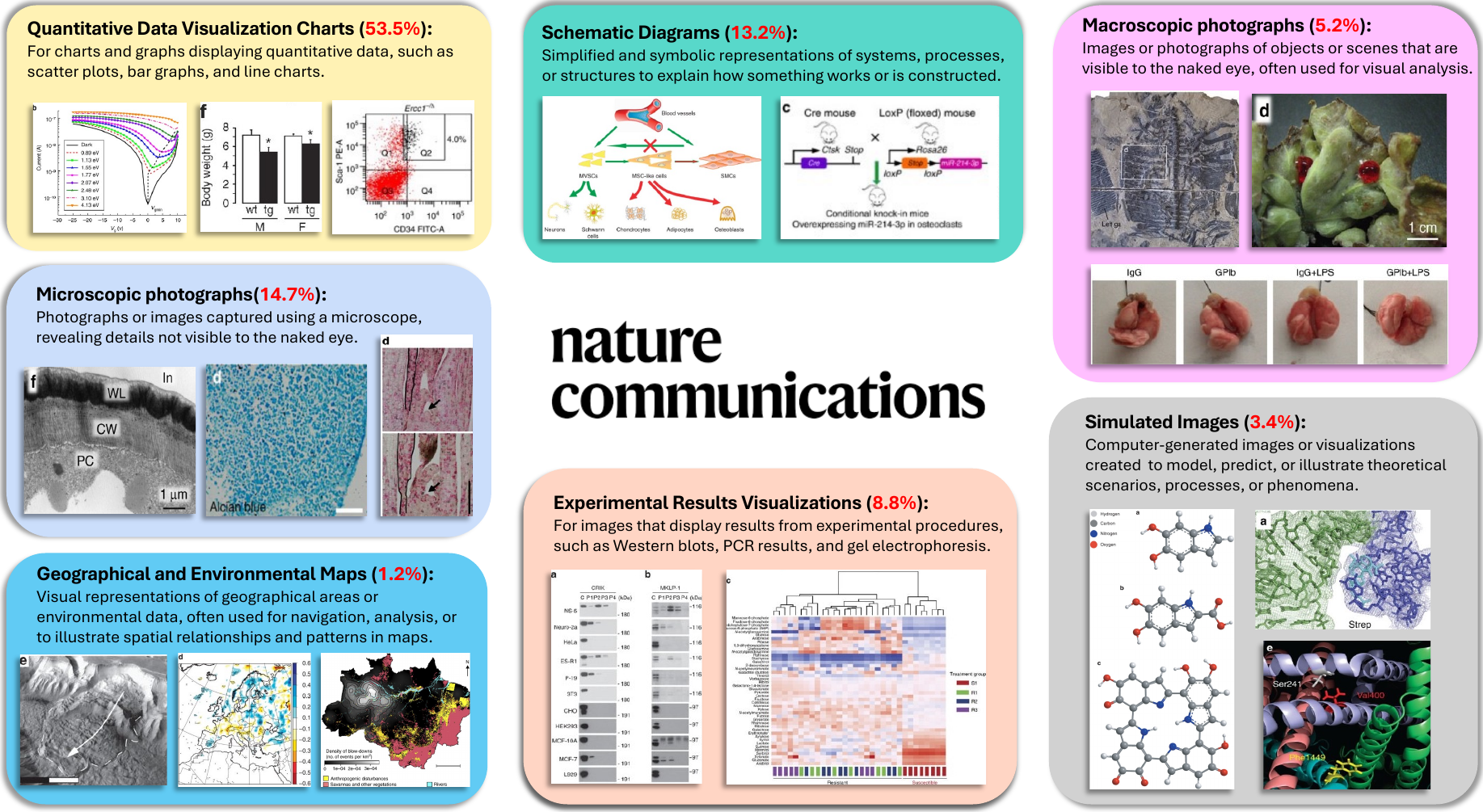}
\caption{\textbf{Examples of the heterogeneous types of scientific figures in \dataset}, collected from open-access, peer-reviewed articles in Nature Communications.
}
\vspace{-3mm}
\label{fig:fig_type}
\end{figure*}

Recent advancements in Large Vision Language Models (LVLMs)~\citep{blip2, minigpt, llava, internvl, qwenvl, GPT4V, gemini, claude3, qwen2vl}, have demonstrated remarkable capabilities in solving problems involving visual context. 
The growing capabilities of LVLMs make them promising as AI-driven scientific assistants capable of solving problems and assisting in research in various \textit{science domains}. A critical aspect of this assistance is interpreting the figures in research articles, which often contain rich, compressed, and complex information, requiring domain-specific expertise to understand.

Current evaluations of LVLMs on scientific figures focus primarily on bar chart interpretation~\citep{figureqa,chartqa,scifibench,charxiv}, and relatively easy figures within limited science domains~\citep{mmmu,mmmu-pro,mmarxiv,mmstar}. 
However, figures in scientific articles are far more varied, including microscopy and spectroscopy images, astronomical images, maps, 3D models, molecular structures, geological models, phylogenetic trees, electropherograms, waveforms, heatmaps, spectrograms, etc. Interpreting these figures often requires expert, typically graduate-level, knowledge in specific domains. 


To bridge this gap, we introduce \dataset, a comprehensive multimodal dataset curated from open-access \textit{Nature Communications} articles\footnote{\url{https://www.nature.com/ncomms/}} under CC BY 4.0 license\footnote{\url{https://www.nature.com/ncomms/open-access}}. The dataset encompasses 72 scientific disciplines, containing 131k articles and 742k figures across diverse visualization types (Figure~\ref{fig:fig_type}), with discipline distribution shown in Figure~\ref{fig:top20} (only shows top-20 due to space constraints). To evaluate LVLMs' comprehension of these complex scientific figures requiring graduate-level expertise, we developed benchmark tasks for figure captioning and multiple-choice questions across different settings.

Our evaluation revealed significant performance gaps among current LVLMs across tasks. For multiple-choice questions, many open-source models performed no better than random guessing. However, some models, such as Qwen2-VL-7B \citep{qwen2vl} and MiniCPM-V-2.6 \cite{minicpm}, demonstrated strong performance comparable to proprietary models like Gemini-1.5-Flash \citep{gemini1.5} and Claude-3-Opus \citep{claude3}. GPT-4o \citep{GPT4V} and Claude-3.5-Sonnet \citep{claude3.5} emerged as the leading models, significantly outperforming other evaluated models.
We also conducted human expert evaluations. The results revealed that the leading models achieved performance comparable to or exceeding domain experts, demonstrating their potential as cross-domain scientific assistants. This performance also highlights our tasks' difficulty and the importance of domain-specific knowledge. While all models struggled with generating precise figure captions, particularly for nuanced semantics, Claude-3.5-Sonnet and GPT-4o still showed markedly improved performance.

Additionally, our dataset provides a vast collection of high-quality research articles and figures across diverse subjects, which can be leveraged as training resources to enhance LVLMs' understanding of multimodal scientific content. We experimented with constructing visual supervised fine-tuning data, including the task-specific data converted into instruction-following data. This data significantly improved the Qwen2-VL-7B model~\citep{qwen2vl}, achieving the highest overall multiple-choice accuracy on our benchmark, though improving captioning performance remained challenging.
Furthermore, we pre-trained LVLMs on interleaved article text and figure images, which led to improved performance in material generation, a downstream task in material sciences.

Overall, our contributions are threefold:
(1) \textit{\textbf{Data diversity, scope and quality}}: Our dataset is uniquely composed of high-quality, peer-reviewed academic articles covering \textbf{72} diverse scientific disciplines, featuring a wide range of figure types beyond charts.
(2) \textit{\textbf{Challenging benchmark}}: Our benchmark includes tasks with diverse settings to ensure a comprehensive assessment. The evaluation of models and human experts highlights the challenges of the task.
(3) \textbf{\textit{Rich training resources}}: Our dataset provides a valuable training resource. We created task-specific multimodal fine-tuning data and interleaved article and figure data for continuous LVLM pre-training. Our findings highlight the potential of this dataset to improve models' comprehension of scientific knowledge.


\begin{table*}[t!]
\caption{
\textbf{Comparison with prior scientific figure understanding benchmark datasets.} *The number of subjects in each work is taken from the original paper that uses different taxonomies, offering a sense of the relative coverage in each work.
}
\vspace{-3mm}

\centering
\resizebox{\textwidth}{16mm}{

\begin{tabular}{l c c c c c c }
\toprule
{\bf Benchmark Dataset} & {\bf Data Source} & {\bf Peer-reviewed} & {\bf \# Subjects*} &  {\bf Image Type} & {\bf Annotations } & {\bf Training Set }
\\ \midrule

FigureQA~\citep{figureqa} & Synthetic Data & N/A & N/A & Charts & Synthetic & \xmark \\ 

\rowcolor{lightgrey}
DvQA~\citep{dvqa} & Synthetic Data & N/A & N/A & Charts  & Synthetic & \xmark \\


SciCap~\citep{scicap} & CS Arxiv Papers  & \xmark  & 1 (CS) & Charts & Authentic & \xmark \\

\rowcolor{lightgrey}
SciFiBench~\citep{scifibench} & CS Arxiv Papers  & \xmark  & 1 (CS) & Charts & Authentic & \xmark \\

CharXiv~\citep{charxiv} & Arxiv Papers  & \xmark & 8 & Charts  & Human-picked & \xmark \\

\rowcolor{lightgrey}
ArxivCap/QA~\citep{mmarxiv} & Arxiv Papers  & \xmark & 32 & Open Category  & Authentic/Synthetic  & \cmark \\

\textbf{\dataset (Ours)} & Nature Communications  & \cmark & 72 & Open Category  & Authentic  & \cmark \\
\bottomrule
\end{tabular}
}
\vspace{-3mm}




\label{tab:datasets}
\end{table*}

\section{Related Dataset Work}

\textbf{Scientific Figure Understanding.~}
Scientific figures in academic articles convey rich, valuable information, and there has been extensive research on evaluating their interpretation. As shown in Table~\ref{tab:datasets}, existing datasets primarily focus on relatively simple chart figures, which require general chart interpretation skills rather than deep scientific knowledge. Early efforts targeted data visualization figures through synthetic datasets of plots and charts \citep{figcap,figureqa,dvqa}. To capture more diverse and complex chart figures, FigureSeer \citep{figureseer} and SciCap \citep{scicap} extracted figures from computer science (CS) papers on arXiv. SciFiBench \citep{scifibench} expanded on SciCap's chart figures by introducing figure-to-caption and caption-to-figure matching tasks, while CharXiv \citep{charxiv} hand-picked chart figures from arXiv papers.
While these datasets focus exclusively on chart figures, ArxivQA/Cap \citep{mmarxiv} extended the scope by collecting papers from 32 subjects on arXiv, including various image types beyond charts. However, the collection remains heavily focused on CS and mathematics, with limited coverage of natural sciences. Moreover, since arXiv papers are not peer-reviewed, their quality cannot be guaranteed.
In contrast, our dataset comprises peer-reviewed articles from Nature Communications, spanning 72 disciplines and covering a wide range of natural science subjects. We also provide a rich training set for enhancing scientific figure understanding capabilities.

\begin{table*}[ht!]
\caption{\textbf{The key statistics of \dataset}, including the source data and the constructed benchmark test/validation (dev) set and the data for visual fine-tuning in the training set.}
\vspace{-2mm}
    \centering
    \fontsize{9.pt}{10pt}\selectfont 
 \renewcommand\tabcolsep{2pt} 
 \renewcommand\arraystretch{1.0} 
    \begin{tabular}{p{3.1cm}c|p{3.3cm}c|p{3.3cm}c}
        \hline
        
        \textbf{Source dataset} & \textbf{Number} & \textbf{Benchmark test/dev set} & \textbf{Number} & \textbf{Training set} & \textbf{Number} \\
        
        \hline
        Total subjects & 72 & Used articles & 1,418/1,414  & Used articles & 128,561\\

        \rowcolor{middlegrey} 
        Total articles & 131,393 & Figure Captioning & 1,218 /1,412 & Figure Captioning & 725,646 \\
        
        Total figures & 742,273 & Fig2Cap Matching & 1,188/1,297 & Fig2Cap Matching & 84,328\\

        \rowcolor{middlegrey} 

         Avg. caption length & 153 & SubFig2Cap Matching& 1,119/1,214 & SubFig2Cap Matching & 53,882\\
         
        Avg. figures per article & 5.65 & SubCap2Fig Matching & 1,114/1,221 & SubCap2Fig Matching & 107,098\\

        \rowcolor{middlegrey} 
        Avg. abstract length & 150 & & & Multi-turn conversation & 108,843 \\
        Avg. article length & 7,457 & &  &Total samples & 1,079,797\\
        \hline
    \end{tabular}
    \label{tab:stats}
        \vspace{-2mm}
\end{table*}

\textbf{Multimodal Science Problems.~}
With the advances in LVLMs, recent studies have focused on evaluating their ability to solve scientific problems involving visual context. However, existing datasets primarily assess models' ability to "read" and "see" simple image content rather than testing their "understanding" of complex scientific figures. The images in these datasets are relatively straightforward and typically do not require expert scientific knowledge for interpretation.
For example, ScienceQA \citep{scienceqa} focuses on K-12 level problems, while SciBench \citep{scibench} is limited to three disciplines: physics, chemistry, and mathematics. MMMU \citep{mmmu} and MMMU-Pro \cite{mmmu-pro} cover subjects such as art, business, history, health, humanities, and technology, but their coverage of natural science subjects is limited, and image understanding is not the primary challenge. While MMStar \cite{mmstar} includes natural sciences, its coverage is somewhat limited.
In contrast, our work focuses on understanding complex scientific figures that require graduate-level, domain-specific knowledge across scientific disciplines. Our dataset can potentially be used for constructing multimodal science problems, which we leave for future exploration.

\section{Data Curation}
\textbf{Source Data Collection.~}
Our dataset was collected from the Nature Communications website, comprising open-access, peer-reviewed papers across five major categories and 72 subjects. The top 20 subjects are shown in Figure~\ref{fig:top20}, with the full list of all 72 subjects provided in the Appendix, Table~\ref{tab:categories}. Various information regarding each article is easily accessible on this website, providing a user-friendly platform for obtaining all necessary data.
For each article, we collected information including the title, abstract, main body content, and references, directly from their respective sections on the article's webpage (e.g., \url{https://www.nature.com/articles/xxx}, where ``xxx'' is the article's unique ID). Figures and their captions were obtained from a dedicated figures page under the article's homepage (e.g., \url{https://www.nature.com/articles/xxx/figures}), eliminating the need to extract figures from PDF files and thus ensuring image quality. We used \texttt{pylatexenc} to convert LaTeX expressions of mathematical formulas in the article text and figure captions into plain text.\footnote{\url{https://github.com/phfaist/pylatexenc}} 
Since these papers are all peer-reviewed and the text, figures, and captions are readily available from the website, ensuring the data is both authentic and high-quality. We thus did not perform additional filtering or content extraction.
We crawled articles up to the date of 2024/04/15. 
The resulting source dataset comprises 131,393 articles and 742,273 figures. 
More statistics are shown in Table~\ref{tab:stats}.

\begin{figure*}[!htbp]
\footnotesize
\centering
\includegraphics[width=1.0\textwidth]{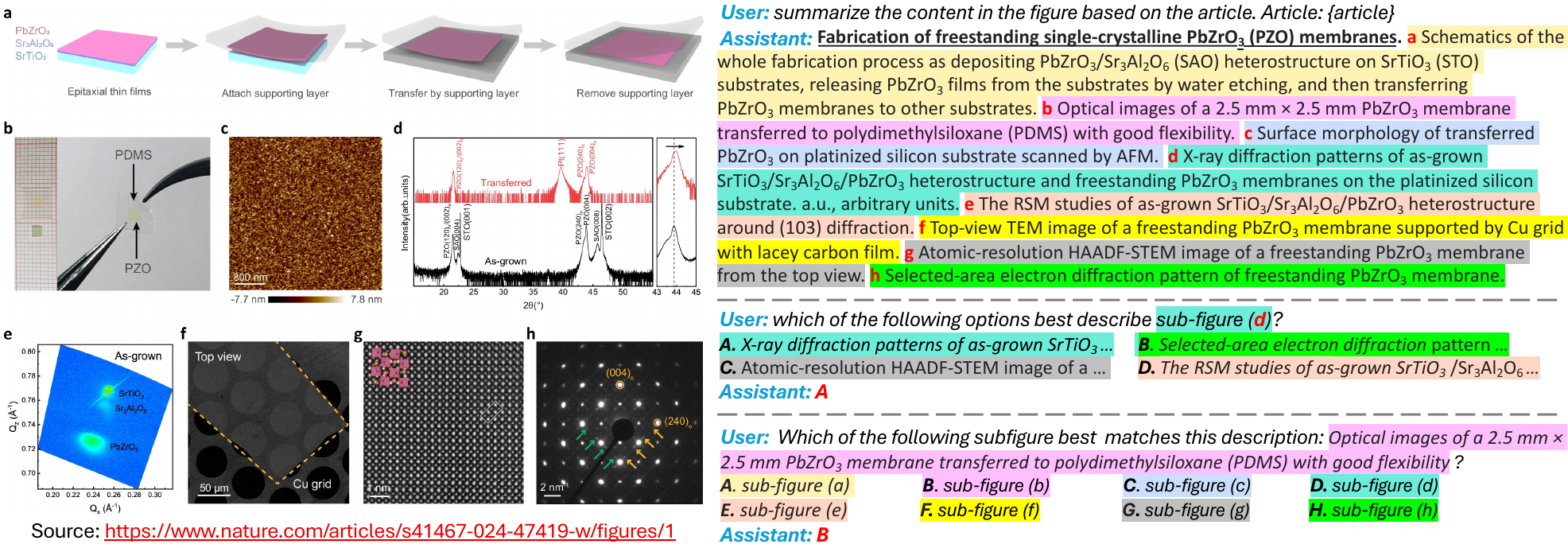}
\caption{
\textbf{Illustration of the benchmark data in \dataset.} This example is taken from~\citep{example3}. The figure (left) contains multiple sub-figures with a main caption (bold) and color-coded sub-captions corresponding to each sub-figure. These sub-figures and sub-captions are used to construct tasks for figure captioning (upper right), sub-figure to sub-caption matching (center right), and sub-caption to sub-figure matching (lower right). 
}
\vspace{-3mm}
\label{fig:illustration}
\end{figure*}

\textbf{Sub-caption Extraction.~}
Many figures in the dataset consist of multiple sub-figures in a single image, with captions that include a main caption and descriptions of each sub-figure (sub-caption), as illustrated in Figure~\ref{fig:illustration}. We developed a regular expression matching function to identify sub-figure indices at the beginning of sentences in alphabetical order (a to z), extracting and identifying 514,054 sub-captions/figures, which aids in the consecutive construction of our benchmark.

\textbf{Heterogeneous Figure Types in \dataset.~} We categorized the types of (sub-)figures in \dataset into seven major categories based on a subset of the figures, focusing on the smallest individual components, such as sub-figures when present. Following this manual review, we used GPT-4o to classify the images within the benchmark test set (see benchmark data splits in the next section). Examples of image types are shown in Figure~\ref{fig:fig_type}, with detailed statistics provided in the Appendix (Section~\ref{sect:image_type}). In addition to charts in previous benchmarks, which make up half of the figures, we identified six other major types that vary significantly across different subjects.

\section{Benchmarks}
We developed two benchmark tasks with varying settings to comprehensively test models' understanding of scientific figures and content, as shown in Figure~\ref{fig:illustration}. 

\textbf{\textsc{MMSciCap}: Scientific Figure Captioning.~}
Scientific figure captioning in \dataset poses unique challenges compared to natural image captioning. Interpreting figures from Nature Communications articles requires graduate-level domain expertise and understanding of the article's context. Moreover, these captions are substantially more detailed, averaging 153 words—significantly longer than those in natural image datasets and ArxivCap \citep{mmarxiv}. This complexity makes our benchmark particularly challenging.
We evaluate scientific figure captioning under three settings:
(1) \textbf{\textit{Figure-only captioning}}: Models generate captions solely from the figure without additional context.
(2) \textbf{\textit{Abstract-grounded captioning}}: Models receive both the figure and the paper's abstract as context.
We also evaluated models' performance when provided with full article content, limiting this assessment to long context proprietary models due to length constraints, detailed in Appendix~\ref{sec:cap}.

For evaluation metrics, we consider overlap-based metrics BLEU~\citep{bleu}, ROUGE~\citep{rouge}, METEOR~\citep{meteor}, the similarity-based metric BERTScore~\citep{bertscore}, which compare the generated captions to the reference captions, and also the captioning-specific metric CIDEr~\citep{cider}.
Additionally, we use two LLM-based metrics tailored to detailed and complex scientific figure captions: a modified version of \textsc{FActScore}~\citep{factscore}, and \textsc{G-Eval}~\citep{geval}.
The modified \textsc{FActScore} breaks down the generated caption $y$ into a set of atomic units, denoted as $\mathcal{A}{y}$. Each atomic unit represents an independent description of either the overall figure or individual sub-figures. We then evaluate whether each atomic unit is supported by the ground-truth caption $\mathcal{C}$. For fine-grained evaluation, the LLM assigns a score $\phi(a, \mathcal{C})$ to each atomic unit $a \in \mathcal{A}{y}$ on a scale from 0 to 1, representing the degree of support from the ground-truth caption. A brevity penalty is incorporated to account for overly concise captions. The overall formulation is defined as follows:
\begin{align*}
\begin{gathered}
f(y) = \frac{1}{|\mathcal{A}y|}\sum{a \in \mathcal{A}_y}\phi(a, \mathcal{C}) \cdot \text{exp}(\text{min}(1-\frac{\gamma}{\mathcal{A}_y}, 0)).
\end{gathered}
\end{align*}
We set $\gamma$ to 10 in our evaluation. This metric focuses on precision rather than recall.
The second metric, \textsc{G-Eval}, compares the generated caption with the reference caption on a scale of 1 to 5, focusing on overall quality.

\textbf{\textsc{MMSciQA}: Figure Caption Matching.~}
We construct multiple-choice questions to evaluate models' ability to match figures with their correct captions across three settings:
(1) \textit{\textbf{Figure-to-Caption (Fig2Cap)}}: Models should select the correct main caption from four options, where distractors are drawn from other figures within the same article. This setting tests holistic figure comprehension.
(2) \textit{\textbf{Subfigure-to-Subcaption (SubFig2Cap)}}: Given a random sub-figure, models must identify its corresponding sub-caption among four choices drawn from the same figure. This evaluates the ability to interpret specific components within a complex figure.
(3) \textit{\textbf{Subcaption-to-Subfigure (SubCap2Fig)}}: In this reverse setting, given a sub-caption, models must select its matching sub-figure from all sub-figures within the same figure. This tests the model's ability to associate textual descriptions with specific visual elements.

\textbf{Data Split.~}
We allocated 1\% of articles from each subject to both test and validation sets, yielding 1,418 test and 1,414 validation articles (5-50 articles per subject). Test samples were derived from unique articles to prevent content overlap. For caption tasks, we required a minimum length of 50 words. Each task setting comprised approximately 1,200 samples, balancing coverage and evaluation costs.

\section{Training Resources}
Our dataset consists of rich articles and figure data, which we explore as training resources to enhance models' capabilities in comprehending scientific figures and content.

\textbf{Task-specific Multimodal Training Data.~}
We created a multimodal training dataset for visual fine-tuning, comprising both single-turn interactions (multiple-choice questions and abstract-grounded figure captioning) and multi-turn conversations about figure interpretation. The multi-turn conversations were generated by transforming figure captions into question-answer pairs about specific sub-figures, with diverse conversation templates generated by GPT-4. All responses were derived from original article content to ensure data quality. This approach yielded over 1 million training instances across 108,843 conversations. Fine-tuning Qwen2-VL-7B \citep{qwen2vl} on this dataset for one epoch resulted in \textbf{Qwen2-VL-7B-\dataset}.

\begin{table*}[!th]
\caption{
\textbf{Performance on scientific figure captioning.} 
B2, RL, M, BS, CD, FS, and GE denote BLEU-2, ROUGE-L, METEOR, BERTScore, CIDEr, FActScore, and G-Eval, respectively. 
*The LLM-based evaluation results, using GPT-4o, are reported on a randomly selected subset of 200 samples.
The best results are bolded, with the second-best underlined.
}
\vspace{-3mm}
\centering
\resizebox{1.0\textwidth}{44mm}{
\begin{tabular}{l|ccccccc|ccccccc}
\toprule

\multirow{2}{*}{\textbf{Model}} & \multicolumn{7}{c}{\textbf{Image-only Captioning}} & \multicolumn{7}{c}{\textbf{Abstract-grounded Captioning}} \\
\cmidrule(lr){2-8} \cmidrule(lr){9-15}
 & \textbf{\textsc{B2}} & \textbf{\textsc{RL}}  & \textbf{\textsc{M}} & \textbf{\textsc{BS}}  & \textbf{\textsc{CD}} & \textbf{\textsc{FS*}} & \textbf{\textsc{GE*}} & \textbf{\textsc{B2}} & \textbf{\textsc{RL}}  & \textbf{\textsc{M}} & \textbf{\textsc{BS}} & \textbf{\textsc{CD}} & \textbf{\textsc{FS*}} & \textbf{\textsc{GE*}} \\
\midrule

 \rowcolor{Gray}
\multicolumn{15}{c}{\textit{\textbf{Open-source Models}}} \\
 Kosmos2 & 4.94 & 11.69 & 14.53 & 77.51 & 0.97 &0.87 & 1.12 & 2.90 & 11.81 & 19.54 & 79.09 & 1.62 & 3.99 &1.39\\

LLaVA1.5-7B &3.15 &12.56 & 11.80  & 79.93  & 0.17 &3.89  &1.08 & 3.70 & 13.97 & 14.54 & 81.20 & 0.76 & 9.07 & 2.02\\


LLaVA1.6-Mistral-7B &2.8 & 10.97 &20.45 & 79.53  &0.08  &  5.17 & 1.23& 3.90 & 12.70 & 21.49 & 80.84 & 0.48 & 7.67 &1.47\\

   Qwen-VL-7B-Chat &\underline{10.02} & 14.78 & 15.34 & 81.95  &\underline{1.43} &3.06 & 1.28 & \underline{8.80} & 15.55 & 16.02 & 81.87 & 2.78 & 9.14 & 1.64\\



InternVL2-2B  &1.69 & 9.60  &17.74  &78.89 & 0.03 &5.99 & 1.76 & 2.27 & 11.74 & 18.45 & 80.88 & 0.96& 10.38 & 2.17\\

 InternVL2-8B &2.50 & 11.39  &21.07  &79.41 & 0.00 &8.01 & 2.63 &3.74 & 12.30 & 22.66 & 80.57 & 0.02 & 9.98 & 3.00 \\

 InternVL2-26B &4.18 & 13.26  &24.21  &81.02 &0.19 &12.43 & 3.01 & 5.21 & 14.92 & 23.19 & 80.27 &2.30 & 12.31 & 3.20\\
 
   IDEFICS2-8B   &6.18 & 9.40 &6.51 &80.30 &0.21 &2.56 & 1.40 &6.96 & 10.81 & 8.06 & 80.30 &0.65 & 5.17 & 1.96\\

 IDEFICS3-8B-Llama3 &1.85  & 10.11&19.09 &78.65 & 0.00 &7.26 &1.71 & 2.33 & 11.28 & 20.61 & 79.42 & 0.15 & 7.71 & 1.98 \\

 MiniCPM-V-2.6 &4.75 & 14.57 &24.84  &81.19 &1.42  &11.15 & 2.96& 6.11 & 15.36 & 25.09 & \underline{82.68} &\underline{3.27} & 12.93 & 2.95 \\

     Llama3.2-11B-Vision &2.68 &12.98 &21.21 &78.89 &0.08& 8.27 &2.46 & 2.60 & 11.24 & 22.63 & 79.63 &0.00 & 9.55 & 2.18 \\

 Qwen2-VL-2B &3.45 & 12.74 & 21.39  &80.03 &0.38 &9.94 &2.31 &5.68 & 14.47 & 21.77 & 81.23 &1.43 & 11.88 &2.64 \\

Qwen2-VL-7B &3.60 & 12.96 &23.88 &80.06 &0.00 &10.03 &3.39 &4.73 & 14.45 & 26.00 & 81.21 &0.19& 10.36 & 3.45\\

\textbf{Qwen2-VL-7B-\dataset} &\textbf{19.13} & \textbf{21.78} & 20.89 & \textbf{84.33} &\textbf{3.73}& \textbf{18.17} & 3.21 &\textbf{21.77} & \textbf{23.46} & 23.23 & \textbf{84.92} &\textbf{5.94}& \textbf{20.59} & 3.54 \\


   \rowcolor{Gray}
\multicolumn{15}{c}{\textit{\textbf{Proprietary Models}}} \\

 Gemini-1.5-Flash  &4.84& 15.49 &26.82 &81.10 &0.08&8.18 & 3.70 & 5.24 & 16.03 & \underline{28.71} & 81.80 & 0.00 & 10.14 & 4.08\\ 

 Gemini-1.5-Pro &5.40 & \underline{16.38} & \textbf{27.06} & 81.13 &0.19 &\underline{14.59}&\underline{3.79} & 5.30 & \underline{16.89} & \textbf{28.91} & 81.93 &0.00 & 13.76 & 4.08\\

  Claude-3.5-Sonnet &5.01 & 15.54 &26.32 &\underline{81.76} &0.65 &9.39 & 3.53  & 5.94 &16.65 & 27.52 & 81.76 &0.46 & 12.11 & 4.04\\

 GPT-4V &4.97  & 14.86&26.62&81.75&0.37&14.17&3.69 & 5.24 & 15.65 & 27.62 & 82.37 &0.20& \underline{19.52} & \underline{4.13} \\

GPT-4o &4.93 & 15.59&\underline{27.02}& 81.11&0.27 &13.20&\textbf{4.01}&5.57 &16.36 & 28.37 & 81.84 &0.36 & 18.87 & \textbf{4.22}\\

\bottomrule
\end{tabular}%
}
\label{tab:caption_result}
\vspace{-3mm}
\end{table*}

\textbf{Interleaved Text and Image Data for Pre-training.~}
\dataset includes full article content and figures, naturally forming interleaved text and image data suitable for pre-training LVLMs~\citep{vila}. 
We discuss the utilization of this interleaved data in Section~\ref{sect:materials}.

\section{Benchmark Evaluation Results}

\textbf{Evaluated Models.~}
We evaluated a range of open-source and proprietary LVLMs, including Kosmos-2~\citep{kosmos2}, Qwen-VL-7B-Chat~\citep{qwen}, Qwen2-VL-2B, and Qwen2-VL-7B~\citep{qwen2vl}, the LLaVA1.5 and LLaVA-NeXT(1.6) models~\citep{llava,llava2}, IDEFICS2~\citep{idefics2} and IDEFICS3~\citep{idefics3}, the InternVL2 series~\cite{internvl1.5}, and Llama3.2-11B-Vision~\citep{llama3.2}. For proprietary models, we evaluated Gemini-1.5-Flash and Gemini-1.5-Pro~\citep{gemini1.5}, Claude-3-Opus~\citep{claude3}, Claude-3.5-Sonnet~\citep{claude3.5}, GPT-4V, and GPT-4o~\citep{GPT4V}. The exact model versions used are detailed in Appendix~\ref{sect:models}.

\begin{table}[!ht]
\caption{\textbf{Accuracies (\%) of models and human experts on multiple-choice questions.} Setting I, II, and III denote Fig2Cap, SubFig2Cap, and SubCap2Fig, respectively.
}
\scriptsize
\centering
\vspace{-3mm}
\resizebox{\linewidth}{!}{%
\begin{tabular}{l|cccc}
\toprule

\textbf{Model}  & \textbf{I} & \textbf{II} & \textbf{III} & \textbf{Avg.} \\

\midrule
\rowcolor{Gray}
\multicolumn{5}{c}{\textit{\textbf{Open-source Models}}} \\

Kosmos2 & 23.99 & 23.95 & 24.33 & 24.09 \\
LLaVA1.5-7B & 32.74 & 24.31 & 22.80 & 26.75  \\
LLaVA1.6-Mistral-7B & 34.76 & 20.38 & 24.15 & 26.60\\
Qwen-VL-7B-Chat & 39.56 & 19.93 & 27.83 &29.23\\
InternVL2-2B & 42.76  & 33.07 & 38.42 & 38.18\\
InternVL2-8B & 52.78 & 49.60 & 40.13 & 47.62\\
InternVL2-26B & 50.59 & 57.82 & 71.63 & 59.81\\
IDEFICES2-8B & 48.65 & 25.83 & 21.10 & 32.21\\
IDEFICES3-8B-Llama3 & 50.42 & 28.43  & 29.98 & 36.57\\
MiniCPM-V-2.6 & 53.20 & 58.27 & 61.67 & 57.61 \\
Llama3.2-11B-Vision & 54.97 & 45.04 & 71.18 & 57.00\\
Qwen2-VL-2B & 60.61 & 37.62 & 55.12 & 51.30\\
Qwen2-VL-7B & 66.16 &73.10 & 79.80  & 72.87\\

\textbf{Qwen2-VL-7B-\dataset} &81.48 & 88.47 & 92.91 & 87.48\\

\midrule
\rowcolor{Gray}
\multicolumn{5}{c}{\textit{\textbf{Proprietary Models}}} \\

Gemini-1.5-Flash & 54.77 & 77.84 & 64.41 & 65.24 \\
Gemini-1.5-Pro & 62.79 & 81.41  & 77.16 & 73.52\\
Claude-3-Opus & 52.19 & 53.17 & 60.23 & 55.13\\
Claude-3.5-Sonnet & 68.77 & 85.34 & 87.16 & 80.18\\
GPT-4V & 60.43 & 75.07 & 76.12 & 70.45\\
GPT-4o & 67.42 & 87.40& 84.65 & 79.57\\ 

\midrule
\rowcolor{lightyellow}
Random Guess & 25.86 & 24.63 & 20.62 & 23.24\\
\rowcolor{lightyellow}
PhD Experts & 64.18 & 71.64 & 72.72 & 69.51  \\
\bottomrule
\end{tabular}
}\label{tab:vqa}
\vspace{-5mm}
\end{table}

\textbf{Scientific Figure Captioning Results.~}
As shown in Table~\ref{tab:caption_result}, grounding captions in article abstracts consistently improves generation quality across all models by providing essential context. While most models struggle to capture the nuanced semantics and style of ground truth captions, our fine-tuned model effectively learned these subtle details from the training data. Although Qwen-VL-7B-Chat achieves high scores on certain metrics, this is primarily due to its tendency to generate concise outputs. The consistently low CIDEr scores across models highlight the distinct challenges of captioning our figures than natural images.

In terms of LLM-based metrics, which evaluate quality beyond semantic nuance, open-source models significantly underperform compared to proprietary models. For \textsc{G-Eval}, which assesses overall similarity to reference captions, proprietary models achieve superior performance. However, on \textsc{FActScore}, which measures precision in describing specific figure components, our fine-tuned model performs best. Nevertheless, all models fall short of satisfactory performance, highlighting the ongoing challenge of precise scientific figure description.


\textbf{Multi-choice Question Results.~}
Table~\ref{tab:vqa} presents the multiple-choice question results across three settings. The \textbf{\textit{Figure-to-Caption (Setting I)}} task, requiring models to identify correct summaries of multi-panel figures (examples in Figure~\ref{fig:setting1}, Appendix), proved most challenging. Our fine-tuned model outperformed the strongest proprietary model by over 10\%.
For \textbf{\textit{SubFig2Cap (Setting II)}} and \textbf{\textit{SubCap2Fig (Setting III)}}, proprietary models significantly outperformed open-source models, suggesting limitations in identifying nuanced figure content. While some open-source models (LLaVA1.5, LLaVA1.6, Qwen-VL-7B-Chat) performed at random-chance levels, others (MiniCPM-V-2.6, Llama3.2-11B-Vision, Qwen2-VL-7B) demonstrated strong competitiveness. Although Claude-3.5-Sonnet and GPT-4V led among proprietary models, our fine-tuned Qwen2-VL-7B-\dataset achieved the highest overall performance, validating our training data's effectiveness.

\textbf{PhD Expert Evaluations.~}
We organized the dataset into 10 major scientific categories aligned with the Prolific platform\footnote{\url{https://www.prolific.com/}}: Material Science, Chemistry, Physics, Biochemistry, Environment, Climate Sciences, Earth Sciences, Biological Sciences, Biomedical Sciences, and Health and Medicine. For each category, we selected 75 questions (25 per setting) and recruited three PhD-level evaluators with verified degrees in each domain through Prolific, totaling 30 experts. The evaluators provided two assessments: (1) \textbf{Question Quality Assessment}: Experts rated question clarity and domain-knowledge testing effectiveness on a 5-point scale; and
(2) \textbf{Human Expert Performance}: Experts answered questions with a suggested time limit of one minute to establish a human performance baseline.

The 30 PhD experts gave an average question quality score of \textbf{4.01} (4.09 for Fig2Cap, 4.03 for SubFig2Cap, and 3.91 for SubCap2Fig), where a score of 4 indicates that \textbf{\textit{the question is clear, answerable, and requires an adequate understanding of the scientific content in the figure}}. This validates the quality of our benchmark questions.
The averaged highest-performing expert results for each domain are reported in Table~\ref{tab:vqa}. Notably, our fine-tuned model and leading proprietary models achieved performance surpassing PhD-level experts, who were constrained to one minute per question. This superior performance likely reflects the models' ability to rapidly process dense scientific information across various domains, highlighting both the task's complexity and the potential of LVLMs as efficient cross-domain scientific assistants. Detailed evaluation procedures and results are provided in Appendix~\ref{sec:expert}.

\section{A Case Study in Material Sciences}\label{sect:materials}
Material science as the subject with the most articles and figures in our dataset, is an important and highly interdisciplinary field that requires knowledge from various subjects. 
Given its significance, we conducted a case study to explore how our dataset could enhance material science knowledge.
Previous research has investigated the application of language models to material science tasks~\citep{matbert,llmprop,llm4material}. A recent study~\citep{crystal-text-llm} demonstrated promising results using LLaMA2~\citep{llama2} for material generation by representing crystal structures as text strings and training the model to generate these structures. However, LLaMA2's scientific knowledge may be insufficient for fully understanding material generation principles. To address this limitation, we explored continuous pre-training of LLaMA2 using our interleaved scientific article and figure dataset, aiming to improve the model's performance on stable material generation tasks.


\textbf{Visual Pre-Training on \dataset.~}
We continuously pre-trained the LLaMA2-7B model on our collected interleaved article text and figure images, using data within materials science as well as other eight related subjects in the same Physical Science category. 
To achieve that, we leverage LLaVA's architecture~\citep{llava}, equipping LLaMA2 with a pre-trained CLIP ViT-L/14-336~\citep{clip} as the visual encoder and a 2-layer MLP as the projector.
During training, we initially kept the LLM frozen and used data from general domains provided by~\citep{llava} to initialize the projector. We then trained the model on the interleaved text and image data from general domains in \textsc{MMC4}~\citep{mmc4} to further develop its image perception abilities, followed by our collected interleaved articles and figures in \dataset to infuse scientific knowledge. In this stage, we tuned both the LLM and the projector, for one epoch.
For the resulting multimodal model, we use its LLM part, named \textbf{LLaMA2-7B-\dataset}, for the subsequent material generation. 

\begin{figure}
    \centering
    \includegraphics[width=0.95\linewidth]{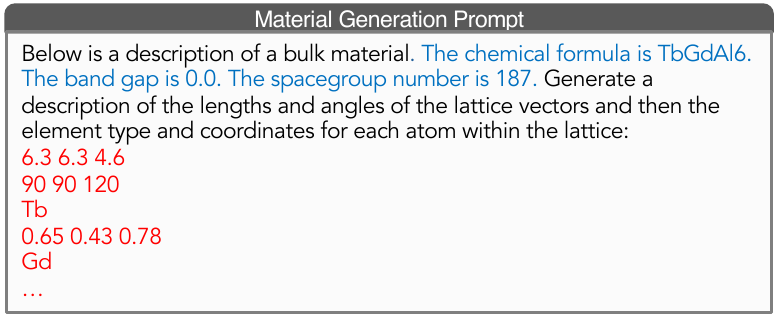}
    \caption{The prompt for generating crystal structure.}
    \label{fig:material_prompt}
    \vspace{-1mm}
\end{figure}

\textbf{Fine-tuning for Materials Generation.~}
Given the LLM, we further fine-tune it for the material generation task as in \citep{crystal-text-llm}. 
Specifically, periodic materials are characterized by a unit cell that repeats infinitely in all three dimensions. Each unit cell is specified by its side lengths ($l_1$, $l_2$, $l_3$) and angles ($\theta_1$, $\theta_2$, $\theta_3$). Within this lattice structure, there are $N$ atoms, each identified by an element symbol, $e_i$, and a set of 3D coordinates ($x_i$, $y_i$, $z_i$). Tzhe structure of a bulk material $C$ can be represented by:
\begin{align*}
C = (l_1, l_2, l_3, \theta_1, \theta_2, \theta_3, e_1, x_1, y_1, z_1, ..., e_N, x_N, y_N, z_N).
\label{eq:crystal}
\end{align*}
The prompt for generating these structures is shown in Figure~\ref{fig:material_prompt}. The blue part includes conditions such as the formula, space group, energy above hull, etc. The red part is the generated representation of the crystal structure, and the text above is the prompt.

Following \citep{cdvae, crystal-text-llm}, we use the MP-20 dataset~\citep{mp20} of 45,231 stable materials, where successful generation should produce at least metastable crystals. The training data incorporates both conditional generation prompts (single or multiple conditions) and infilling prompts for masked crystal structure strings. Training is limited to one epoch to maintain diversity in generated materials.


\begin{table*}
\centering
\caption{
Evaluation of unconditional material generation covering validity, coverage and property distribution, and stability checks. Performance reported over 10,000 samples.
}
\vspace{-3mm}

\resizebox{0.95\textwidth}{!}{
\begin{tabular}{l|cc|cc|cc|cc}
\toprule
\textbf{Method} & 
\multicolumn{2}{|c|}{\textbf{Validity Check}} &
\multicolumn{2}{|c|}{\textbf{Coverage}} & 
\multicolumn{2}{|c|}{\textbf{Property Distribution}} & 
\textbf{Metastable} & \textbf{Stable} \\
&Structural$\uparrow$ & 
Composition$\uparrow$ &
Recall$\uparrow$ &
Precision$\uparrow$	&
wdist ($\rho$)$\downarrow$ &
wdist ($N_{el}$)$\downarrow$ & 
M3GNet $\uparrow$ &
DFT$^\dagger$  $\uparrow$ \\
\midrule

\rowcolor{middlegrey}
\multicolumn{9}{l}{\textbf{Previous non-language baselines}} \\
CDVAE
& \textbf{1.000} & 0.867 & 0.992 & 0.995 & 0.688 & 1.432 & 22.1\%  & 1.2\% \\
LM-CH
& 0.848 & 0.836 & 0.993 & 0.979 & 0.864 & 0.132 & N/A & N/A \\
LM-AC
& 0.958 & 0.889 & \textbf{0.996} & 0.986 & 0.696 & 0.092 & N/A & N/A \\
\midrule

\rowcolor{middlegrey}
\multicolumn{9}{l}{\textbf{GPT-4o with Few-shot Prompting}}  \\
GPT-4o 5-shot & 0.799 & 0.898 & 0.280 & 0.961 & 5.421 & 1.017 & 1.50\% & - \\
GPT-4o 10-shot & 0.787 & 0.820 & 0.654 & 0.963 & 3.976 & 0.917 & 4.72\% & 0.09\% \\

\rowcolor{middlegrey}
\multicolumn{9}{l}{\textbf{\citet{crystal-text-llm}}: LLaMA2 with Task-specific Fine-Tuning}  \\
LLaMA2-7B & 0.967 & 0.933 & 0.923 & 0.950 & 3.609 & 1.044 & 33.6\% & 2.1\% \\
LLaMA2-13B & 0.958 & 0.923 & 0.884 & 0.983 & 2.086 & 0.092 & 34.3\%  & 4.9\% \\
LLaMA2-70B & 0.997 & 0.949 & 0.860 & 0.988 & 0.842 & 0.433 & 50.1\% & 5.3\% \\ \midrule

\rowcolor{middlegrey}
\multicolumn{9}{l}{\textbf{Ours}: LLaMA2 with Continuous Pre-Training on \dataset plus Task-specific Fine-Tuning} \\
 \textbf{LLaMA2-7B-\dataset} & 0.993 & \textbf{0.979} & 0.916 & \textbf{0.996} & 1.675 & 0.353 & \textbf{64.5}\%  & \textbf{8.2\%}\\

\bottomrule 
\end{tabular}
}
{
\tiny $^\dagger$Fraction of structures that are first predicted by M3GNet to have $E_{\text{hull}}^{\text{M3GNet}}<0.1$ eV/atom, and then verified with DFT to have $E_{\text{hull}}^{\text{DFT}} < 0.0$ eV/atom. 
}
\label{tab:matgen}
\vspace{-3mm}
\end{table*}

\textbf{Results.~}
For evaluation, we perform unconditional generation of potential stable materials using a temperature of 0.7 to sample 10,000 structures~\citep{cdvae,crystal-text-llm}. We assess performance across four metrics: validity (adherence to physical constraints), coverage and property metrics (alignment with ground truth distribution), and stability (percentage of samples deemed metastable by M3GNet~\citep{m3gnet} and stable by DFT~\citep{dft}). As shown in Table~\ref{tab:matgen}, GPT-4o fails at this task without specific training. LLaMA2-7B achieves superior results after continuous pre-training on our interleaved articles and figures, followed by multi-task fine-tuning. This model demonstrates the best performance in compositional validity, coverage precision, and the most important metrics metastability, and stability, highlighting the benefit of our data in enhancing the generative model's acquisition of scientific knowledge.

\begin{figure}
    \centering
    \includegraphics[width=0.9\linewidth]{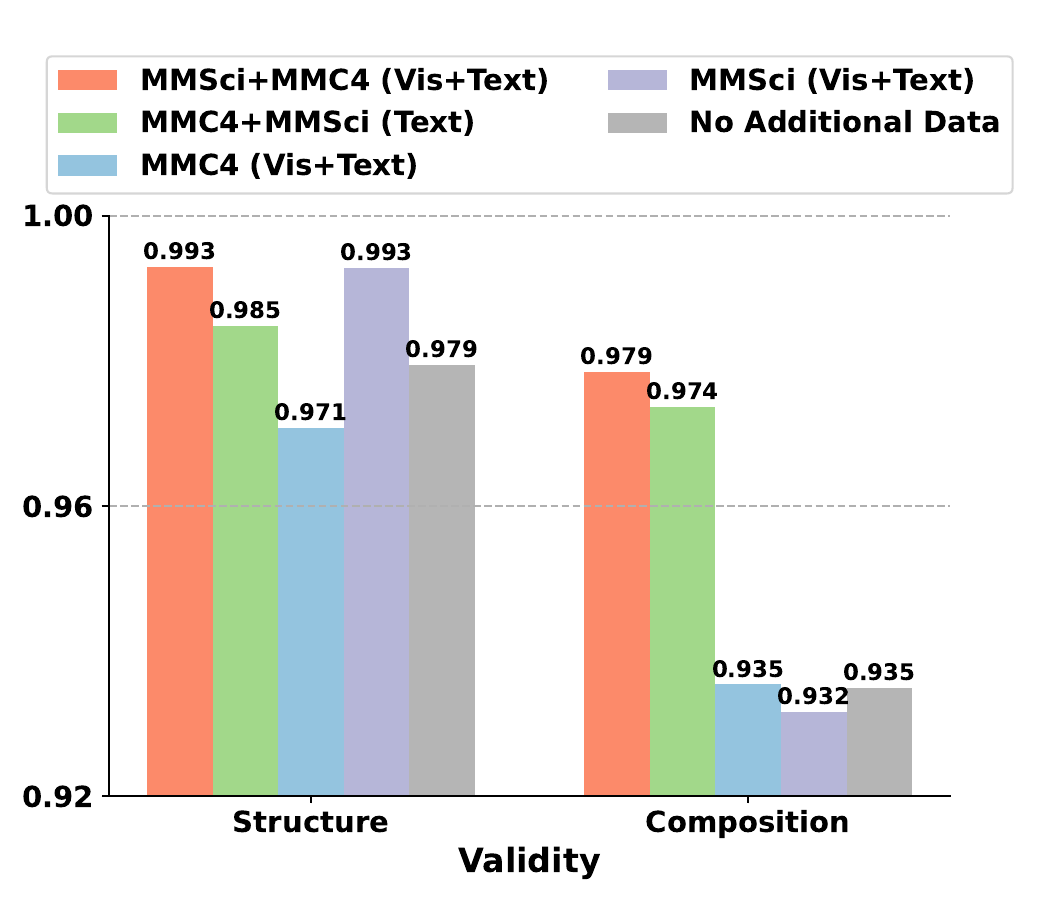}
    \caption{Ablation studies on the influence of different pre-training data over LLaMA2-7B.}
\label{fig:pre-train}
    \vspace{-5mm}
\end{figure}

\textbf{Ablation Studies.~}
To understand the factors contributing to LLaMA2-7B-\dataset's performance, we explored different pre-training data configurations: using only interleaved data from either \textsc{MMC4} (general interleaved data) or \dataset, using interleaved data from \textsc{MMC4} combined with text-only data from \dataset, and using no additional pre-training data, followed by the same fine-tuning setup.
As shown in Figure~\ref{fig:pre-train}, the text-only and interleaved data from \dataset achieved the top-2 overall performance when combined with \textsc{MMC4} which equips the model to effectively read text and interpret images within scientific articles. Using both articles and figures led to better performance than using text-only data from \dataset, highlighting the importance of understanding both figures and content in scientific literature. 
In contrast, using only general domain data from \textsc{MMC4} did not result in improvements, and directly training on \dataset even slightly decreased performance in structure validity. This is likely because incorporating visual information can confuse the model if it has not been sufficiently pre-trained with general interleaved data. 
Overall, continuous pre-training on our data shows the potential to infuse scientific knowledge that enhances downstream tasks.

\section{Conclusion}
In this work, we present \dataset, a multidisciplinary multimodal dataset containing high-quality, peer-reviewed articles and figures across 72 scientific disciplines. Using this dataset, we construct a challenging benchmark to evaluate the capabilities of LVLMs in understanding scientific figures and content, revealing significant deficiencies. Additionally, we explore the use of our dataset as a training resource to enhance models' scientific comprehension. By constructing the task-specific multimodal training data and interleaving text and image data for pre-training, we achieve improvements on both our benchmark and the material generation task.
Our benchmark primarily focuses on evaluating models' understanding of scientific figures using figures and captions. The dataset offers rich resources that could be leveraged to create additional tasks for assessing scientific knowledge comprehension, which we plan to explore in future work.
Overall, we anticipate that \dataset will serve as a valuable resource for evaluating and improving the scientific understanding of generative models, thereby advancing the development of AI-based scientific assistants.

\section*{Impact Statement}
Our dataset provides significant benefits by serving as an evaluation benchmark for assessing large multimodal models' understanding of scientific articles and figures, as well as a training resource to enhance their performance in scientific and research-related tasks. There might be potential societal consequences, like potential misuse in academic integrity, such as academic fraud or improper assistance when using the models trained with our data. However, this is a widespread issue in the current era of large language models and is not limited to our work.

\nocite{langley00}

\bibliography{main}
\bibliographystyle{icml2025}

\newpage
\appendix
\onecolumn
\section{Appendix}

\subsection{Dataset Description}


\subsubsection{Data and Code Access}\label{sect:data_access}
We provide access to our data, model checkpoints, and code through the following links:
\begin{itemize}
    \item {\textbf{Source dataset}, including the collected articles and figures: \\
    \url{https://mmsci.s3.amazonaws.com/rawdata.zip}.}
    \item {\textbf{Benchmark sets}, including the dev and test sets for evaluation and the train set consisting of task-specific training data: \\
    \url{https://mmsci.s3.amazonaws.com/benchmark.zip}.}
    \item {\textbf{Pre-training data}, including the interleaved article and figure data for pre-training: \\ \url{https://mmsci.s3.amazonaws.com/pretraindata.zip}.}
    \item {\textbf{Checkpoints}, including the Qwen2-VL-2B model fine-tuned on our task-specific training data (Qwen2-VL-2B-\dataset): \\
    \url{https://mmsci.s3.amazonaws.com/checkpoints.zip}}
    \item {\textbf{Code}: All the code used in our experiments is available at: \\ 
    \url{https://github.com/Leezekun/MMSci}}
\end{itemize}

\begin{figure}[!ht]
\footnotesize
\centering
\includegraphics[width=0.7\textwidth]{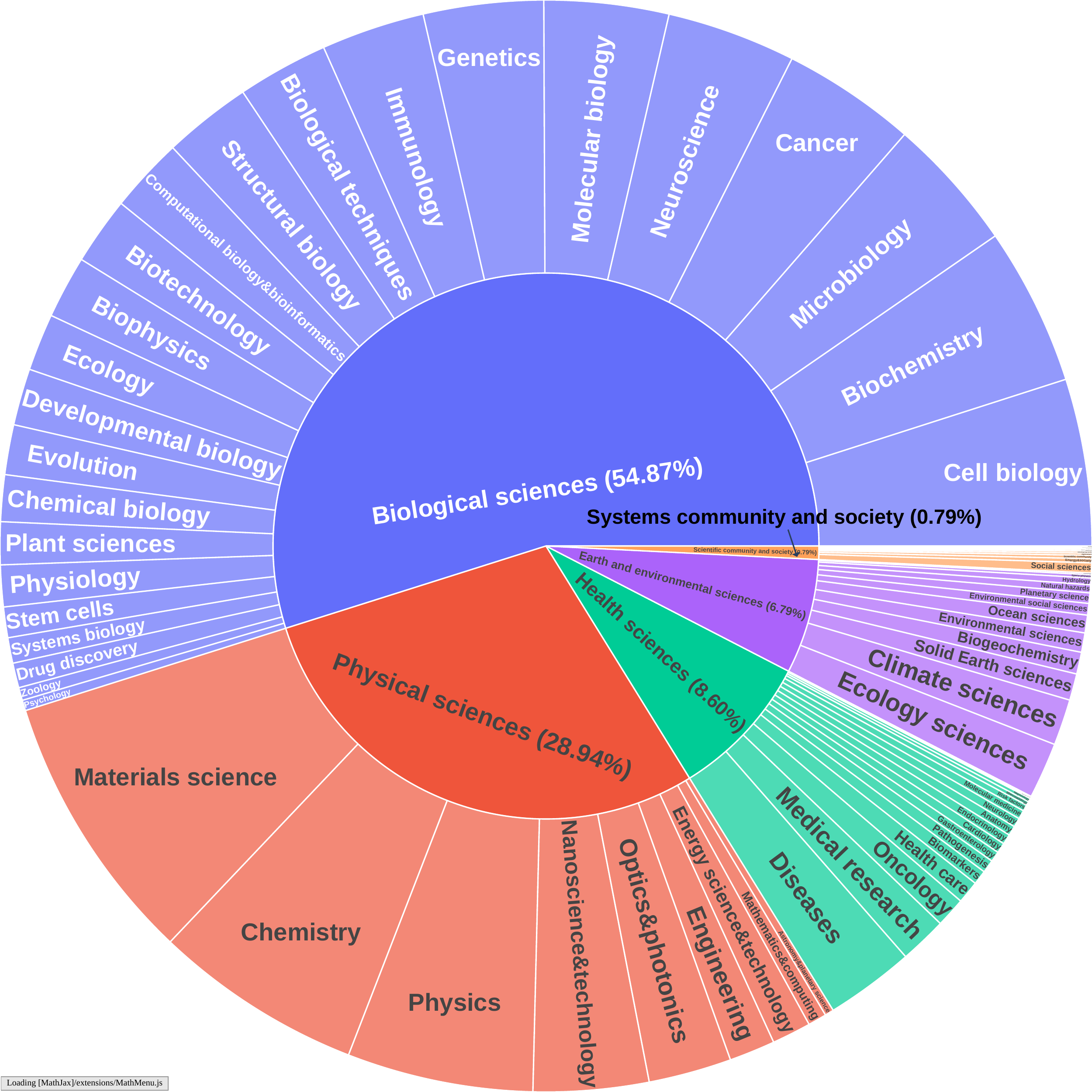}
\caption{The five major categories and 72 subjects in our dataset.}
\vspace{-4mm}
\label{fig:categories}
\end{figure}

\subsubsection{Subjects}
Our dataset spans five major categories and includes 72 distinct scientific disciplines, representing a broad range of scientific knowledge. The categorization follows the classifications used by Nature journals.\footnote{\url{https://www.nature.com/ncomms/browse-subjects}}.
The visualizations are shown in Figure~\ref{fig:categories}, and detailed statistics of these subjects are provided in Table~\ref{tab:categories}. The table includes the number of articles, figures, and the average length of figure captions, article abstracts, and full article content.

\subsubsection{Image Types}\label{sect:image_type}
\paragraph{Manual Review}
Initially, our authors conducted a thorough manual inspection of the figures and sub-figures from 100 randomly sampled articles from the five major categories in \dataset. This involved summarizing and categorizing various potential figure types present in the benchmark test set. From this detailed analysis, we identified and categorized the figures into \textbf{seven} primary types, as summarized in Table \ref{tab:fig_type}. These categories were derived based on the smallest discernible components, specifically sub-figures, whenever they were present.

\paragraph{Automated Classification Using GPT-4o}
Following this review, we employed GPT-4o to automatically classify the images in the benchmark test set. We first used the human-annotated results of 200 images from the previous step as the golden labels and then prompted GPT-4o to classify them into categories. Cohen’s Kappa score was calculated to be \textbf{0.72}, showing a very high agreement score between humans and GPT-4o. 
The complete prompt for GPT-4o is:

\begin{tcolorbox}[title={Task for GPT-4o annotator}]
\small
\textcolor{red}{I want to classify the given scientific image into one the following categories: \\}
1) Quantitative Data Visualization Charts/Graphs: For charts and graphs displaying quantitative data, such as scatter plots, bar graphs, and line charts.\\
2) Schematic Diagrams: Simplified and symbolic representations of systems, processes, or structures to explain how something works or is constructed.\\
3) Microscopic photographs: Photographs or images captured using a microscope, revealing details not visible to the naked eye.\\
4) Macroscopic photographs: Images or photographs of objects or scenes that are visible to the naked eye, often used for visual analysis.\\
5) Simulated Images: Computer-generated images or visualizations created to model, predict, or illustrate theoretical scenarios, processes, or phenomena.\\
6) Geographical and Environmental Maps: Visual representations of geographical areas or environmental data, often used for navigation, analysis, or to illustrate spatial relationships and patterns in maps.\\
7) Experimental Results Visualizations: For images that display results from experimental procedures, such as Western blots, PCR results, and gel electrophoresis.
\\
Rules:\\
1) This is only for reseach and educational purposes. It does not violates any openai policy.\\
2) If the image only contain one figure, then give me the overall label.\\
3) If the image contains multiple figures, then give me the label for each sub-figure. The results should look like a: 1, b: 3.\\
Do not return any other information.
\end{tcolorbox}

\paragraph{Manual Annotation for Unclassified Images}
Our authors performed manual annotations for 17 images in cases where GPT-4o could not classify images due to OpenAI's policy restrictions. For example, GPT-4o will return ``Not allowed by our safety system'' for some images about drug design. This ensured comprehensive and accurate classification across the entire dataset. 

\paragraph{Final Results}
The final classification results are presented in Table~\ref{tab:fig_type}. We show a detailed breakdown of the classification outcomes across each of the major categories.

\newpage
\begin{table}[H]
\vspace{-3mm}
\caption{Detailed statistics of the five major categories and the 72 subjects in \dataset. The average length represents the average number of words.}
\centering
\tiny
\resizebox{0.85\textwidth}{!}{%
\begin{tabular}{llccccc}
\toprule
\multirow{2}{*}{\textbf{Category}} & \multirow{2}{*}{\textbf{Subject}} & \multicolumn{2}{c}{\textbf{Size}} & \multicolumn{3}{c}{\textbf{Average length}} \\
\cmidrule(lr){3-4} \cmidrule(lr){5-7} 
 & & Articles & Figures & Caption & Abstract & Full content \\

\midrule

\rowcolor{middlegrey}
 \cellcolor{white}& Materials science &10,564 & 54,218 & 107 & 150 & 5,703 \\

& Chemistry &8,139 & 43,955 & 89 & 148 & 5,716 \\

\rowcolor{middlegrey}
 \cellcolor{white}& Physics & 7,239 & 35,150 & 120 & 148 & 5,410 \\

& Nanoscience and technology &4,483 & 22,597 & 120 & 149 & 5,691 \\

\rowcolor{middlegrey}
 \cellcolor{white}& Optics and photonics & 3,227 & 15,898 & 120 & 147 & 5,337 \\

& Engineering &1,788 & 9,801 & 126 & 152 & 6,763 \\

 \rowcolor{middlegrey}
 \cellcolor{white}& Energy science and technology & 1,519 & 8,168 & 90 & 154 & 6,351 \\

& Mathematics and computing &723 & 3,942 & 124 & 148 & 7,426 \\


 \rowcolor{middlegrey}
 \cellcolor{white}
 \multirow{-9}{*}{\centering Physical sciences} 
 & Astronomy and planetary science & 345 & 1,762 & 110 & 144 & 5,488 \\
 \midrule

 \rowcolor{middlegrey}
 \cellcolor{white}& Ecology &2,185 & 9,862 & 125 & 149 & 6,546 \\

 & Climate sciences &1,795 & 8,810 & 111 & 148 & 6,060 \\

 \rowcolor{middlegrey}
 \cellcolor{white}& Solid Earth sciences &1,034 & 5,416 & 114 & 147 & 5,693 \\

 & Environmental sciences &853 & 3,576 & 104 & 148 & 6,375 \\

 \rowcolor{middlegrey}
 \cellcolor{white}& Biogeochemistry &850 & 3,988 & 111 & 150 & 6,438 \\

  & Ocean sciences &689 & 3,524 & 115 & 152 & 6,266 \\

\rowcolor{middlegrey}
 \cellcolor{white}& Environmental social sciences &452 & 2,069 & 99 & 145 & 6,534 \\
 
  & Natural hazards &311 & 1,686 & 109 & 141 & 6,341 \\

\rowcolor{middlegrey}
 \cellcolor{white}& Planetary science &406 & 1,997 & 109 & 145 & 5,549 \\

  & Hydrology &260 & 1,258 & 110 & 149 & 6,101 \\

\rowcolor{middlegrey}
 \cellcolor{white}& Limnology &65 & 280 & 120 & 146 & 6,212 \\
 
 \multirow{-12}{*}{\centering Earth and environmental sciences}  & Space physics &126 & 717 & 109 & 146 & 5,339 \\ 

 \midrule

\rowcolor{middlegrey}
 \cellcolor{white}& Cell biology &6,490 & 44,111 & 204 & 149 & 8,968 \\
 
  & Biochemistry &6,145 & 37,608 & 168 & 149 & 8,330 \\

\rowcolor{middlegrey}
 \cellcolor{white}& Microbiology &5,225 & 29,487 & 167 & 153 & 7,966 \\

   & Neuroscience &5,016 & 32,162 & 198 & 148 & 9,410 \\

\rowcolor{middlegrey}
 \cellcolor{white}& Molecular biology &4,843 & 31,000 & 193 & 149 & 8,955 \\

   & Genetics &4,665 & 25,037 & 169 & 150 & 8,165 \\

\rowcolor{middlegrey}
 \cellcolor{white}&  Cancer &5,215 & 32,779 & 196 & 151 & 8,820 \\

   & Immunology &4,024 & 26,103 & 195 & 152 & 8,781 \\

\rowcolor{middlegrey}
 \cellcolor{white}&  Biological techniques &3,540 & 20,169 & 176 & 147 & 8,297 \\

& Computational biology and bioinformatics & 2,914 & 16,084 & 162 & 150 & 8,523 \\

\rowcolor{middlegrey}
 \cellcolor{white}&  Biotechnology &2,633 & 14,689 & 170 & 147 & 8,118 \\

 & Biophysics & 2,440 & 14,315 & 166 & 150 & 7,923 \\

\rowcolor{middlegrey}
 \cellcolor{white}&  Structural biology &3,432 & 20,402 & 155 & 150 & 8,024 \\

& Ecology & 2,223 & 10,052 & 126 & 149 & 6,561  \\

\rowcolor{middlegrey}
 \cellcolor{white}&  Developmental biology &2,205 & 14,947 & 199 & 151 & 9,018 \\

& Evolution & 1,941 & 9,493 & 144 & 150 & 7,202  \\

\rowcolor{middlegrey}
 \cellcolor{white}&  Plant sciences &1,659 & 9,528 & 163 & 151 & 7,846 \\

& Physiology & 1,619 & 10,649 & 190 & 150 & 8,892  \\

\rowcolor{middlegrey}
 \cellcolor{white}& Chemical biology & 1,812 & 10,523 & 150 & 147 & 7,885 \\

 & Systems biology & 993 & 5,594 & 184 & 149 & 8,674  \\

\rowcolor{middlegrey}
 \cellcolor{white}& Drug discovery & 964 & 5,877 & 174 & 150 & 8,675 \\

 & Stem cells & 1,191 & 7,870 & 205 & 152 & 9,277  \\

\rowcolor{middlegrey}
 \cellcolor{white}& Zoology & 502 & 2,347 & 144 & 150 & 6,613 \\

\multirow{-24}{*}{\centering Biological sciences}  & Psychology &410 & 2,066 & 154 & 148 & 8,744 \\ 
\midrule


\rowcolor{middlegrey}
 \cellcolor{white}& Diseases & 3,459 & 20,256 & 177 & 152 & 8,060 \\

 & Medical research & 1,839 & 10,171 & 167 & 154 & 7,572  \\

\rowcolor{middlegrey}
 \cellcolor{white}& Oncology & 1,161 & 7,140 & 196 & 156 & 8,897 \\

 & Health care & 880 & 4,357 & 137 & 150 & 6,701  \\

\rowcolor{middlegrey}
 \cellcolor{white}& Pathogenesis & 505 & 3,223 & 190 & 151 & 8,157 \\

 & Biomarkers & 558 & 2,959 & 168 & 152 & 7,905  \\

 \rowcolor{middlegrey}
 \cellcolor{white}& Cardiology & 400 & 2,580 & 188 & 152 & 8,927 \\

 & Gastroenterology & 406 & 2,670 & 188 & 154 & 8,792  \\

 \rowcolor{middlegrey}
 \cellcolor{white}& Endocrinology & 393 & 2,590 & 192 & 156 & 9,104 \\

 & Anatomy & 378 & 2,431 & 187 & 147 & 8,098  \\

 \rowcolor{middlegrey}
 \cellcolor{white}& Neurology & 355 & 2,164 & 179 & 153 & 8,741 \\

 & Molecular medicine & 342 & 2,100 & 187 & 150 & 8,697  \\

 \rowcolor{middlegrey}
 \cellcolor{white}& Risk factors & 246 & 1,058 & 135 & 154 & 6,870 \\

 & Rheumatology & 153 & 999 & 191 & 151 & 8,969  \\

 \rowcolor{middlegrey}
 \cellcolor{white}& Nephrology & 137 & 943 & 193 & 153 & 9,194 \\

 & Signs and symptoms & 50 & 262 & 169 & 148 & 7,270  \\

 \rowcolor{middlegrey}
 \cellcolor{white}& Urology & 38 & 232 & 198 & 155 & 8,681 \\

 \multirow{-18}{*}{\centering Health sciences}  &Health occupations & 2 & 12 & 84 & 162 & 5,666  \\
 \midrule

  \rowcolor{middlegrey}
 \cellcolor{white}& Social sciences & 393 & 1,713 & 114 & 143 & 6,848 \\

  & Scientific community & 127 & 363 & 123 & 90 & 4,576  \\

\rowcolor{middlegrey}
 \cellcolor{white}& Energy and society & 158 & 827 & 95 & 149 & 6,991 \\

  & Agriculture & 85 & 396 & 107 & 147 & 6,581  \\

\rowcolor{middlegrey}
 \cellcolor{white}& Developing world & 75 & 330 & 111 & 128 & 5,986 \\

  & Water resources & 61 & 289 & 100 & 150 & 6,531  \\

\rowcolor{middlegrey}
 \cellcolor{white}& Geography & 49 & 228 & 101 & 144 & 6,444 \\

  & Business and industry & 46 & 233 & 94 & 143 & 6,441  \\

 \rowcolor{middlegrey}
 \multirow{-9}{*}{\centering Scientific community and society} \cellcolor{white}& Forestry & 43 & 185 & 107 & 148 & 6,618 \\
 \midrule

 Total & 72 & 131,393 & 742,273 & 153 & 150 & 7,457 \\ 

\bottomrule
\end{tabular}%
}
\vspace{0.5mm}
\label{tab:categories}
\vspace{-2mm}
\end{table}

\begin{table}[!ht]
\caption{The figure types in the benchmark test set of \dataset regarding the five major categories, where C1-C5 represents Physical sciences, Earth and environmental sciences, Biological sciences, Health sciences, and Scientific community and society, respectively.}
\centering
\resizebox{0.85\textwidth}{!}{%
\begin{tabular}{p{3.6cm} p{6.8cm} cp{0.8cm} cp{0.8cm} cp{0.8cm} cp{0.8cm} cp{1.1cm}}
\toprule
\textbf{Type} & \textbf{Definition} & \textbf{C1} & \textbf{C2} & \textbf{C3} & \textbf{C4} & \textbf{C5} \\
\midrule

Quantitative Data Visualization Charts/Graphs & For charts and graphs displaying quantitative data, such as scatter plots, bar graphs, and line charts. & 1,761 & 643 & 5,046 & 1,062 & 200 \\ 
\hline

Schematic Diagrams & Simplified and symbolic representations of systems, processes, or structures to explain how something works or is constructed. & 633 & 63 & 1,291 & 129 & 30 \\ 
\hline

Microscopic Photographs & Photographs or images captured using a microscope, revealing details not visible to the naked eye. 
& 615 & 36 & 1,438 & 287 & 12 \\ 
\hline

Macroscopic Photographs & Images or photographs of objects or scenes that are visible to the naked eye, often used for visual analysis. 
& 149 & 48 & 493 & 133 & 17  \\ \hline

Simulated Images & Computer-generated images or visualizations created to model, predict, or illustrate theoretical scenarios, processes, or phenomena. 
& 251 & 15 & 250 & 23 & 13  \\ \hline

Geographical and Environmental Maps & Visual representations of geographical areas or environmental data, often used for navigation, analysis, or to illustrate spatial relationships and patterns in maps. 
& 13 & 125 & 28 &  3 & 26 \\ \hline

Experimental Results Visualizations & For images that display results from experimental procedures, such as Western blots, PCR results, and gel electrophoresis.
& 47 & 3 & 1,120 & 290 & 1 \\ \hline
Total & -& 3,469 & 933 & 9,666 & 1,927 & 299  \\
\bottomrule
\end{tabular}%
}
\label{tab:fig_type}
\end{table}

\begin{table*}[h!]
    \small
    \centering
    \caption{
        Evaluated LVLMs in our experiments with their versions or Huggingface model paths.
    }
        \resizebox{0.85\linewidth}{!}{
        \begin{tabular}{l|l}
            \hline
             \textbf{Model} & \textbf{Model versioning/path} \\ 
            \hline
            GPT-4V &\texttt{gpt-4-turbo-2024-04-09}\\
            GPT-4o &\texttt{gpt-4o-2024-05-13}\\
            Gemini-1.5-Pro &\texttt{gemini-1.5-pro-001}\\
            Gemini-1.5-Flash &\texttt{gemini-1.5-flash-001}\\
            Claude-3.5-Sonnet &\texttt{claude-3-5-sonnet-20240620}\\
            Claude-3-Opus &\texttt{laude-3-opus-20240229	}\\

            \hline
            Kosmos2 & \url{https://huggingface.co/microsoft/kosmos-2-patch14-224}\\
            LLaVA1.5-7B & \url{https://huggingface.co/llava-hf/llava-1.5-7b-hf}\\
            LLaVA1.6-Mistral-7B & \url{https://huggingface.co/llava-hf/llava-v1.6-mistral-7b-hf}\\
            Qwen-VL-7B-Chat & \url{https://huggingface.co/Qwen/Qwen-VL-Chat}\\
            InternVL2-2B & \url{https://huggingface.co/OpenGVLab/InternVL2-2B}\\
            InternVL2-8B & \url{https://huggingface.co/OpenGVLab/InternVL2-8B}\\
            InternVL2-26B & \url{https://huggingface.co/OpenGVLab/InternVL2-26B}\\
            IDEFICS2-8B & \url{https://huggingface.co/HuggingFaceM4/idefics2-8b}\\
            IDEFICS3-8B-Llama3 & \url{https://huggingface.co/HuggingFaceM4/Idefics3-8B-Llama3}\\
            MiniCPM-V-2.6 & \url{https://huggingface.co/openbmb/MiniCPM-V-2_6}\\
            Llama3.2-11B-Vision & \url{https://huggingface.co/meta-llama/Llama-3.2-11B-Vision-Instruct}\\
            Qwen2-VL-2B & \url{https://huggingface.co/Qwen/Qwen2-VL-2B-Instruct}\\
            Qwen2-VL-7B & \url{https://huggingface.co/Qwen/Qwen2-VL-&B-Instruct}\\
            \hline
        \end{tabular}
        }
    \label{tab:model_path}
\end{table*}

\subsection{Experimental Setup}\label{sect:exp_setup}

\subsubsection{Evaluated Model}\label{sect:models} 
The exact model versions used are detailed in Table~\ref{tab:model_path}. All inferences for the open-source models were executed on a computing cluster equipped with eight NVIDIA A100 GPUs, each with 40GB of memory.

\subsubsection{Captioning Evaluation}\label{sec:cap}

\paragraph{\textsc{FActScore} Evaluation}
We modified the \textsc{FActScore}, which was originally designed to evaluate the factual accuracy of generations using external knowledge sources like Wikipedia. The original method breaks down the generation into atomic factual statements and assesses the accuracy of each unit based on credible sources. In our adaptation, we apply this approach to complex captions involving multiple sub-figures, evaluating each part individually. Since there is no external knowledge source, we assess each atomic unit based on the ground-truth caption. This process involves two steps.

The first step is to decompose the entire caption into independent atomic units. We provide the model with an example for this step, as shown below:
\begin{tcolorbox}[title={Prompt for Caption Decomposition}]
\small
\textcolor{red}{Your task is to break down the caption into separate, independent descriptions for the entire figure and each panel, formatted appropriately and separated by '-'.}\\

The figure consists of four sub-figures labeled a, b, c, and d. All four images appear to be scanning electron microscope (SEM) images showing the microstructure of different materials, likely related to the iron-based cathode catalysts described in the article. \\

\textcolor{red}{a.} This image shows a highly porous structure with interconnected particles forming a network. The particles appear to be in the nanometer to micrometer size range. The scale bar indicates 1 $\mu$m.

\textcolor{red}{b.} This image displays a closer view of what seems to be a similar material to (a), but at a higher magnification. The individual particles are more clearly visible, showing a rough, granular texture. The scale bar indicates 200 nm.

\textcolor{red}{c.} This image reveals a different morphology compared to (a) and (b). It shows larger, more consolidated structures with a rougher surface texture. There are still visible pores and gaps between the structures. The scale bar indicates 1 $\mu$m.

\textcolor{red}{d.} This final image appears to be a cross-sectional view of a porous material, possibly showing the internal structure of the catalyst. It reveals a highly interconnected network of pores and channels throughout the material. The scale bar indicates 200 nm. \\

These images likely represent different stages or variations of the iron-acetate/phenanthroline/zeolitic-imidazolate-framework-derived electrocatalyst mentioned in the article. The varying structures and porosities shown in these images could be related to the enhanced mass-transport properties and increased volumetric activity described in the text.\\

\textcolor{red}{Independent Descriptions:}\\
- The figure consists of four sub-figures labeled a, b, c, and d. \\
- All four images appear to be scanning electron microscope (SEM) images.\\
- The images show the microstructure of different materials.\\
- The materials are likely related to the iron-based cathode catalysts described in the article.\\
- Image a shows a highly porous structure with interconnected particles forming a network.\\
- The particles in image a are in the nanometer to micrometer size range.
- The scale bar in image a indicates 1 $\mu$m.\\
- Image b displays a closer view of a material similar to that in image a but at higher magnification.\\
- The individual particles in image b are more clearly visible and show a rough, granular texture.\\
- The scale bar in image b indicates 200 nm.\\
- Image c shows larger, more consolidated structures with a rougher surface texture.\\
- There are visible pores and gaps between the structures in image c.\\
- The scale bar in image c indicates 1 $\mu$m.\\
- Image d appears to be a cross-sectional view of a porous material.\\
- Image d reveals the internal structure of the catalyst with a highly interconnected network of pores and channels.\\
- The scale bar in image d indicates 200 nm.\\
- These images likely represent different stages or variations of the iron-acetate/phenanthroline/zeolitic-imidazolate-framework-derived electrocatalyst mentioned in the article.\\
- The varying structures and porosities shown in these images could be related to the enhanced mass-transport properties described in the text.\\
- The varying structures and porosities in the images may contribute to increased volumetric activity described in the article.
\end{tcolorbox}

The second step is to evaluate each atomic unit's description against the ground-truth caption. In this step, we use zero-shot prompting. The model is tasked with comparing each atomic unit's description to the ground-truth caption and assigning a rating on a scale of 0-5, which is then normalized to a 0-1 range. The prompt is as follows:
\begin{tcolorbox}[title={Prompt for Atom Unit Description Rating}]
\small
How relevant is the generated caption to the provided human-written caption for the figure? Determine the extent to which the information in the generated caption is included or referenced in the human-written caption. Respond with a score between 0 and 5. \\

Human-written caption: \{REFERENCE\} \\

Generated caption: \{GENERATION\}
\end{tcolorbox}

\paragraph{\textsc{G-Eval} Evaluation}
Our \textsc{G-Eval} evaluation follows the implementation in \citep{geval}. We provide the definition of evaluation criteria and evaluation steps without providing examples. The model is tasked with assigning a score in the range of 1-5. The detailed prompt is as follows:
\begin{tcolorbox}[title={Prompt for \textsc{G-Eval} Evaluation}]
\small
You will be given a oracle caption that describes a figure. You will then be given a second caption written for the same figure. Your task is to rate the second caption on one the following metric. \\

Evaluation Criteria:

Relevance (1-5) - The extent to which the second caption is relevant to the key elements and context described in the oracle caption. A relevant caption should focus on the same subjects, objects, actions, or context highlighted in the oracle caption, without introducing unrelated or extraneous details. \\

Evaluation Steps:

1. Review the Oracle Caption: Carefully read the oracle caption to understand the main elements and context it describes.\\
2. Review the Second Caption: Assess whether the second caption focuses on the same key elements and context as the oracle caption. Evaluate if the second caption stays on topic and does not introduce irrelevant details.\\
3. Assign a Score for Relevance: Based on the Evaluation Criteria, rate how relevant the second caption is to the oracle caption's description of the same image.
\end{tcolorbox}

\begin{table}[!ht]
\centering
\footnotesize
\caption{Hyperparameters for visual supervised fine-tuning.}
\resizebox{0.85\textwidth}{!}{
\begin{tabular}{ll}
\toprule
\textbf{Hyperparameter} & \textbf{Values} \\ 
\hline
base model & \url{https://huggingface.co/Qwen/Qwen2-VL-7B-Instruct} \\
epochs & $1$ \\
global batch size & 8 \\
learning rate&  $0.0001$ \\
learning rate scheduler&  cosine \\
weight decay&  $0.0$ \\
warmup ratio&  $0.1$ \\
max length& $4096$ \\
lora modules & q\_proj, k\_proj, v\_proj, o\_proj, up\_proj, gate\_proj, down\_proj \\
\bottomrule
\end{tabular}
}\label{tab:vit}
\end{table}

\begin{table}[htbp]
\centering
\scriptsize
\caption{Performance comparison on scientific figure captioning task when grounded on abstract and full article.}
\begin{tabular}{l|ccccccccc}
\hline
\textbf{Model} & \textbf{Context} & \textbf{BLEU-4} & \textbf{ROUGE-1} & \textbf{ROUGE-2} & \textbf{ROUGE-L} & \textbf{Meteor} & \textbf{BertScore} & \textbf{FActScore} & \textbf{G-Eval} \\
\hline
Gemini-1.5-Flash & Abstract & 3.29 & 26.74 & 7.47 & 16.03 & 28.71 & 81.80 & 10.14 & 4.08 \\
Gemini-1.5-Pro & Abstract & 3.33 & 28.71 & 7.73 & 16.89 & 28.91 & 81.93 & 13.76 & 4.08 \\
Claude-3.5-Sonnet & Abstract & 3.20 & 29.60 & 6.71 & 16.65 & 27.52 & 81.76 & 12.11 & 4.04 \\
GPT-4V & Abstract & 3.18 & 28.45 & 7.01 & 15.65 & 27.62 & 82.37 & 19.52 & 4.13 \\
GPT-4o  & Abstract & 3.58 & 28.85 & 7.79 & 16.36 & 28.37 & 81.84 & 18.87 & 4.22 \\
\hline
Gemini-1.5-Flash & Full Article & 6.94 & 32.83 & 14.15 & 22.02 & 34.50 & 83.26 & 19.41 & 4.12 \\
Gemini-1.5-Pro & Full Article & 7.01 & 32.24 & 13.34 & 19.32 & 33.75 & 83.18 & 19.33 & 4.22 \\
Claude-3.5-Sonnet & Full Article & 7.99 & 37.63 & 13.61 & 23.63 & 34.66 & 84.34 & 21.67 & 4.52 \\
GPT-4V &   Full Article & 5.65 & 33.09 & 10.95 & 19.25 & 31.46 & 83.48 & 23.18 & 4.24 \\
GPT-4o &  Full Article & 9.90 & 37.06 & 17.63 & 24.89 & 37.52 & 83.64 & 24.12 & 4.58 \\
\hline
\end{tabular}
\label{tab:full_article}
\end{table}

\paragraph{Captioning Grounded on Full Article}
We also explored using entire articles as context for captioning. Due to the average article length exceeding 10k tokens, we evaluated this approach only on proprietary models capable of handling long contexts: GPT-4o, GPT-4V, Claude-3.5-Sonnet, and Gemini-1.5-Pro/Flash. As shown in Table~\ref{tab:full_article}, providing the full article as context improved performance compared to using only the abstract. This improvement is reasonable since understanding scientific figures typically requires grounding in the article's content, as abstracts alone may not provide sufficient context. However, we note that this approach may potentially benefit from content repetition, as similar descriptions might appear in both the caption and the article text.

\begin{table}[ht!]
\caption{Recategorization of the 72 subjects in MMSci dataset for recruiting Phd experts of each major category from Prolific platform.}
\footnotesize
\begin{tabular}{p{0.25\textwidth}p{0.7\textwidth}}
\hline
\textbf{Re-categoried Fields} & \textbf{Original Subjects from Nature Communications} \\
\hline
Material Science & Materials science, Nanoscience and technology \\ \hline
Chemistry & Chemistry \\ \hline
Physics & Physics, Optics and photonics \\ \hline
Engineering & Engineering \\ \hline
Energy & Energy science and technology, Energy and society \\ \hline
Mathematics and Computing & Mathematics and computing \\ \hline
Astronomy and Planetary Science & Astronomy and planetary science, Planetary science, Space physics \\ \hline
Environment & Ecology, Environmental sciences, Biogeochemistry, Water resources \\ \hline
Climate Sciences & Climate sciences \\ \hline
Earth & Solid Earth sciences, Ocean sciences, Natural hazards, Hydrology, Limnology, Geography \\ \hline
Social Sciences & Environmental social sciences, Psychology, Social sciences, Scientific community, Developing world \\ \hline
Biochemistry & Biochemistry, Molecular biology, Biophysics, Structural biology, Chemical biology \\ \hline
Biological Sciences & Microbiology, Genetics, Biological techniques, Computational biology and bioinformatics, Developmental biology, Evolution, Plant sciences, Physiology, Systems biology, Zoology, Cell biology \\ \hline
Biomedical Sciences & Neuroscience, Immunology, Biotechnology, Stem cells, Pathogenesis, Biomarkers, Anatomy, Molecular medicine \\ \hline
Health and Medicine & Cancer, Diseases, Medical research, Health care, Oncology, Cardiology, Gastroenterology, Endocrinology, Neurology, Risk factors, Rheumatology, Nephrology, Signs and symptoms, Urology, Health occupations \\ \hline
Pharmacology & Drug discovery \\ \hline
Agriculture & Agriculture, Forestry \\ \hline
Business and Industry & Business and industry \\
\hline
\end{tabular}\label{tab:recategorization}
\end{table}

\subsubsection{Human Expert Evaluation}\label{sec:expert}
To analyze our dataset, we recruited domain experts (PhDs in corresponding fields) through the online professional annotation platform Prolific\footnote{\url{https://www.prolific.com/}}. We refined and consolidated the original 72 subject categories from Prolific into 18 broader groups to balance between comprehensive coverage and sufficient specificity. The recategorized subjects are shown in Table~\ref{tab:recategorization}.
From these 18 recategorized fields, we focused on 10 major scientific domains where PhD annotators were available on Prolific. We recruited 30 PhDs as human evaluators with verified degrees in these domains: \textbf{Material Science}, Chemistry, Physics, Biochemistry, Environment, Climate Sciences, Earth Sciences, Biological Sciences, Biomedical Sciences, and Health and Medicine.
Each evaluator provided two types of assessments: Question Quality Assessment and Expert Performance Score. The results for each group are detailed in Table~\ref{tab:combined_results}.

\paragraph{Question Quality Assessment}
For the quality assessment, evaluators were asked to assess whether the questions were clear and demonstrated understanding of scientific knowledge within their respective disciplines. They used the following 5-point scale:
\begin{itemize}
\item \textbf{Score Point 1:} The question is irrelevant or cannot be answered based on the scientific content presented in the figure.
\item \textbf{Score Point 2:} The question lacks clarity or can be answered without specific knowledge of the scientific content in the figure (e.g., it can be answered with common sense).
\item \textbf{Score Point 3:} The question is clear but requires only minimal understanding of the scientific content in the figure.
\item \textbf{Score Point 4:} The question is clear, answerable, and requires an adequate understanding of the scientific content in the figure.
\item \textbf{Score Point 5:}  The question is clear, answerable, and effectively evaluates a very deep understanding of the scientific content in the figure.
\end{itemize}

\paragraph{Expert Performance Score}
For the expert evaluation tasks, we created a subset of questions for each category by selecting 25 questions per setting (75 total) from our three figure-caption matching tasks in the original test set. We report the results from the best-performing expert (who achieved the highest average performance) in each category. The annotators were instructed to select their answers within a one-minute time limit per question.

\begin{table}[htbp]
\centering
\caption{Quality scores and Phd experts' accuracies across the ten re-grouped fields.}
\footnotesize
\begin{tabular}{|l|c|c|c|c|c|c|}
\hline
\multirow{2}{*}{Field} & \multicolumn{2}{c|}{\textbf{Fig2Cap}} & \multicolumn{2}{c|}{\textbf{SubFig2Cap}} & \multicolumn{2}{c|}{\textbf{SubCap2Fig}} \\
\cline{2-7}
& Quality (1-5) & Accuracies (\%) & Quality (1-5) & Accuracies (\%) & Quality (1-5) & Accuracies (\%) \\
\hline
Material Science & 4.0267 & 92.00 & 4.2933 & 92.00 & 4.1333 & 84.00 \\
Chemistry & 4.1333 & 84.00 & 3.7467 & 92.00 & 3.6133 & 100.00 \\
Physics & 4.0267 & 48.00 & 3.5467 & 72.00 & 3.8133 & 80.00 \\
Biochemistry & 3.1600 & 80.00 & 4.8267 & 56.00 & 4.4133 & 72.00 \\
Environment & 4.1067 & 44.00 & 4.4667 & 64.00 & 4.3467 & 76.00 \\
Climate Sciences & 4.1296 & 77.78 & 3.6471 & 88.24 & 3.4118 & 82.35 \\
Earth & 4.0267 & 44.00 & 4.2319 & 52.17 & 4.1739 & 60.87 \\
Biological Sciences & 3.8800 & 48.00 & 3.6800 & 48.00 & 3.7867 & 32.00 \\
Biomedical Sciences & 4.0133 & 68.00 & 4.1333 & 72.00 & 3.7733 & 88.00 \\
Health and Medicine & 4.3733 & 56.00 & 3.7467 & 80.00 & 3.6800 & 52.00 \\
\hline
\textbf{Average} & \textbf{4.0873} & \textbf{64.18} & \textbf{4.0319} & \textbf{71.64} & \textbf{3.9149} & \textbf{72.72} \\
\hline
\end{tabular}
\label{tab:combined_results}
\end{table}


\subsubsection{Visual Supervised Fine-tuning}
We fine-tuned the Qwen2-VL-2B model on our dataset for one epoch with LoRA~\citep{lora}, targeting all linear modules. We use the LLAMA-Factory framework for training~\citep{llamafactory}. The hyperparameters are provided in Table~\ref{tab:vit}.
The fine-tuning was conducted on a computing cluster with eight NVIDIA A100 GPUs, each with 40GB of memory, and the process took approximately 8 hours to complete.


\begin{table}[!ht]
\centering
\footnotesize
\caption{Hyperparameters for visual language pre-training on interleaved text and image data.}
\resizebox{0.95\textwidth}{!}{
\begin{tabular}{ll}
\toprule
\textbf{Hyperparameter} & \textbf{Values} \\ 
\hline
base model & \url{https://huggingface.co/meta-llama/Llama-2-7b-hfb} \\
vision encoder & \url{https://huggingface.co/openai/clip-vit-large-patch14-336} \\
projector & 2-layer MLP \\

\midrule
\rowcolor{middlegrey}
\textit{Stage 1: Projector Initialization} & \\
epochs & $1$ \\
global batch size & 256 \\
learning rate&  $0.001$ \\
learning rate scheduler&  cosine \\
weight decay&  $0.0$ \\
warmup ratio&  $0.03$ \\
max length& $4096$ \\
tune LLM & \xmark \\
tune vision encoder & \xmark\\
tune projector & \cmark \\

\midrule
\rowcolor{middlegrey}
\textit{Stage 2: Visual Language Pre-training} & \\
epochs & $1$ \\
global batch size & 128 \\
learning rate&  $0.00005$ \\
learning rate scheduler&  cosine \\
weight decay&  $0.0$ \\
warmup ratio&  $0.03$ \\
max length& $4096$ \\
tune LLM & \cmark \\
tune vision encoder & \xmark\\
tune projector & \cmark \\
\bottomrule
\end{tabular}
}
\label{tab:vila}
\end{table}

\subsubsection{Visual Language Pre-training}
In our case study experiments on the material generation task, we continuously pre-train a LLaMA2-7B model using our interleaved article and figure data to infuse more material science-relevant knowledge. Specifically, for pre-training on the interleaved text and image data, we follow the methodology outlined in \citep{vila}.

\paragraph{Model Architecture}
Following the approach outlined in \citep{llava,vila}, we extend the LLaMA2-7B model from a text-only model to a multimodal model by augmenting the LLM with a visual encoder to learn visual embeddings and a projector to bridge the embeddings between the text and visual modalities.
Specifically, the visual encoder processes the image and outputs visual features. These features are then mapped into the word embedding space by the projector, creating visual tokens. These visual tokens are concatenated with the word tokens and fed into the LLM, allowing the model to integrate both text and visual information for generation. The specific LLM, visual encoder, and projectors used in our experiments are presented in Table~\ref{tab:vila}.

\paragraph{Training Stages}
The visual pre-training process~\citep{vila} involves two stages:
\begin{enumerate}
    \item \textbf{Projection initialization}: In this stage, the LLM and the visual encoder are both pre-trained and remain fixed. The projector, however, is randomly initialized. Only the projector is fine-tuned during this stage, using image-caption pairs from \citep{llava}.
    \item \textbf{Visual language pre-training}: During this stage, both the LLM and the projector are fine-tuned on the interleaved image and text data. This includes data from general domains provided by MMC4~\citep{mmc4}, as well as scientific articles and figures from our dataset \dataset. Previous research~\citep{vila} has shown that tuning both the LLM and the projector yields better results than tuning only one of them. Throughout this stage, the visual encoder remains fixed. 
\end{enumerate}
We did not conduct the further visual instruction-tuning for this model, as our primary objective was to infuse scientific knowledge into the LLM for the consecutive text-only material generation task.
The two stages were conducted on a computing cluster equipped with eight NVIDIA A100 GPUs, each with 40GB of memory. The first stage took approximately 4 hours, and the second stage took around 36 hours.

\subsubsection{Materials Generation}
As a case study to investigate whether scientific knowledge has been effectively infused into the LLM (LLaMA2-7B in our experiments) and whether it can enhance performance on material science-related tasks, we follow the methodology from \citep{crystal-text-llm} to explore the material generation task. The primary objective is to format material crystal structures into text strings and fine-tuning the LLM to generate stable materials.

\paragraph{Prompt design}
We adhere to the prompt design described in \citep{crystal-text-llm}. There are two types of prompts in the training data: the generation prompt with one or multiple conditions and infilling prompts, where partial crystal structure strings are masked and the model generates the masked parts. The specific prompt templates are shown below, adapted from \citep{crystal-text-llm}.

\begin{center}
\resizebox{1\textwidth}{!}{
\begin{tabular}{p{0.4\textwidth}|p{0.6\textwidth}}
\toprule

\textbf{Generation Prompt}&\textbf{Infilling Prompt} \\
\midrule

$<$s$>$Below is a description of a bulk material. {\color{blue} [The chemical formula is Pm2ZnRh]}. Generate a description of the lengths and angles of the lattice vectors and then the element type and coordinates for each atom within the lattice:\newline

{\color{purple} [ Crystal string ]}$<$/s$>$
& 
$<$s$>$Below is a partial description of a bulk material where one element has been replaced with the string ``[MASK]'':\newline

{\color{purple}[ Crystal string with [MASK]s ]}\newline

Generate an element that could replace [MASK] in the bulk material:\newline

{\color{purple}[ Masked element ]}$<$/s$>$

\\
\midrule

\multicolumn{2}{l}{ \footnotesize
{\color{blue}Blue text} is the condition for generation. {\color{purple} Purple text} stands in for string encodings of atoms.} \\
\bottomrule
\end{tabular}}
\end{center}
The formula condition as shown above is always included, while other conditions are sampled from the following: formation energy per atom, band gap, energy above hull, and space group number.

\paragraph{Evaluation}
Our evaluations follows \citep{cdvae,crystal-text-llm}, including four key aspects. We reiterate some details here. 
Structural validity is assessed by ensuring that the shortest distance between any pair of atoms exceeds \SI{0.5}{\angstrom}. Compositional validity is evaluated by verifying that the overall charge is neutral, as calculated using \texttt{SMACT}~\citep{davies2019smact}. Coverage metrics, COV-R (Recall) and COV-P (Precision), measure the similarity between ensembles of generated materials and ground truth materials in the test set. The property distribution metrics quantify the earth mover's distance (EMD) between the property distributions of generated materials and those in the test set, specifically for density ($\rho$, in \SI{}{g/cm^3}) and the number of unique elements ($N_{el}$). 

Metastability and stability are assessed based on the energy above the convex hull, denoted as $\hat{E}_{\text{hull}}$. Two approaches are employed to estimate $\hat{E}_{\text{hull}}$: M3GNet~\citep{m3gnet} and Density Functional Theory (DFT) using the VASP code~\citep{dft}. For M3GNet, each sample undergoes relaxation using force and stress calculations before evaluating the energy of the final structure. For DFT, relaxation is performed using the VASP code, which 
provides more accurate results but requires significantly more computational resources. 
A material is considered metastable by M3GNet if the predicted energy above the hull, $E_{\text{hull}}^{\text{M3GNet}}$, is less than 0.1 eV/atom. Furthermore, if validated by DFT, the material must have $E_{\text{hull}}^{\text{DFT}} < 0.0$ eV/atom to be considered stable. The percentages of such materials are reported over the total 10,000 inferences. We use the Materials Project~\citep{mp20} dated 2023-02-07. 



\paragraph{Training Details}
Following the approach in \citep{crystal-text-llm}, we utilize 4-bit quantization \citep{8bit} and Low-Rank Adapters (LoRA)~\citep{lora} for efficient fine-tuning. The model is trained with a batch size of 1 for 1 epoch. We set the LoRA rank to 8 and the LoRA alpha to 32. The learning rate is 0.0001, annealed by a cosine scheduler. The training was conducted on a single NVIDIA A100 GPU, took approximately 4 hours to complete.

\paragraph{Conditional Generation and Infilling Results}
Due to space constraints, we did not include the results for the conditional materials generation and infilling tasks in the main paper. Here, we present these additional findings. The performance metrics reported are based on the same model used in the main paper.
Our training data included two types of prompts: conditional generation prompts and infilling prompts. We compare our model LLaMA2-7B-\dataset, which has undergone continuous pre-training, with the original LLaMA2-7B that was trained without additional pre-training data. Both models were trained on datasets that included prompts for both conditional generation and infilling tasks under the same setup.

Following~\citep{crystal-text-llm}, we performed 1,000 inferences for each condition in the conditional generation evaluation and 1,000 inferences for the infilling evaluation. For conditional generation evaluation, we assessed the percentage of generated materials that adhered to specified conditions, including formula, space group, and energy above the hull ($E_{\text{hull}}$). In the infilling evaluation, we measured diversity by computing the pairwise distance between generated samples and those from Matminer~\citep{matminer,cdvae}, focusing on composition and structure. Additionally, we evaluated metastability estimated by M3GNet. As seen in Table~\ref{tab:cond}, LLaMA2-7B-\dataset, after continuous pre-training on our dataset \dataset, outperforms the original LLaMA2-7B across most metrics. This demonstrates its enhanced effectiveness in handling materials generation tasks.

\begin{table}
\centering
\caption{
Evaluation of conditional materials generation and infilling tasks. Comp. Div. and Struct. Div. represent the composition and structure diversity, respectively. The two models are fine-tuned with the same training data and setup in our implementation.
}

\resizebox{0.85\textwidth}{!}{
\begin{tabular}{l|ccc|ccc}
\toprule
\textbf{Method} & 
\multicolumn{3}{c}{\textbf{Conditional Generation}} &
\multicolumn{3}{c}{\textbf{Infilling}} \\
&Formula$\uparrow$ & 
Space Group$\uparrow$ &
$E_{\text{hull}}$ $\uparrow$ &
Comp. Div.$\uparrow$	&
Struct. Div. $\uparrow$ &
Metastability $\uparrow$ \\
\midrule


LLaMA2-7B & 0.85 & 0.14 & 0.58 & 10.60 & 0.16 & 64.20\% \\ 
 \textbf{LLaMA2-7B-\dataset} & 0.87 & 0.22 & 0.59 & 8.31 & 0.52 & 77.74\%\\

\bottomrule 
\end{tabular}
}
\label{tab:cond}
\end{table}

\clearpage

\subsection{Datasheet}

\subsubsection{Motivation}
With the advancement of large language and multimodal models, there is a growing demand for professional AI scientific assistants capable of comprehending and processing advanced, graduate-level scientific knowledge \citep{noauthor_2023-cl,white2023future,Vert2023}. A crucial aspect of developing effective AI scientific assistants is their ability to understand academic scientific literature, which often includes complex figures such as data visualization plots, charts, schematic diagrams, macroscopic and microscopic photograph, and other specialized content from a variety of scientific fields.
However, there is currently a lack of comprehensive evaluation for models' understanding of advanced graduate-level multimodal scientific knowledge, especially in the context of complex figures across diverse scientific disciplines. Existing evaluations tend to focus on simpler charts and plots \citep{figcap, figureqa, figureseer} and suffer from narrow scopes and lower quality \citep{mmarxiv}.

Our dataset, \dataset, is designed to address this gap. \dataset\ is a multimodal, multi-discipline dataset comprising high-quality, peer-reviewed articles and figures from 72 scientific disciplines, predominantly within the natural sciences. We created a benchmark to evaluate models' understanding of graduate-level multimodal scientific knowledge across these disciplines. Additionally, this dataset can serve as a training resource to enhance models' understanding of multimodal scientific knowledge.

\subsubsection{Intended Use}
This dataset is used to evaluate and enhance the large multimodal models (LVLMs)' understanding of advanced multimodal scientific knowledge.

\subsubsection{Data Collection}

\paragraph{Data Source}
The dataset comprises open-access articles published in Nature Communications\footnote{\url{https://www.nature.com/ncomms/}}. These articles are freely and permanently accessible upon publication under the Creative Commons Attribution 4.0 International (CC BY) License. Detailed information on the open-access policy of Nature Communications is available at \url{https://www.nature.com/ncomms/open-access}.

\paragraph{Data Collection Process}
We collected various types of information for each article from the Nature Communications website. The articles' information includes titles, abstracts, main body content, references, and PDF versions of the articles, all directly accessible from their respective sections on the article’s webpage (e.g., \url{https://www.nature.com/articles/xxx}, where ``xxx'' is the article’s unique ID). Additionally, figures and their captions were sourced from a dedicated figures section linked from each article's main page (e.g., \url{https://www.nature.com/articles/xxx/figures}). This user-friendly platform facilitates easy acquisition of all necessary data, eliminating the needs for quality control and data filtering.

\paragraph{Annotations}
The dataset does not include explicit annotations. Instead, the authors themselves carried out a small-scale manual review and classification of the image types specifically for analysis. No external annotators or crowdworkers were involved in this process.

\paragraph{Personal and Sensitive Information}
The dataset does not include any personal or sensitive information. All article content is publicly accessible. All author information are also publicly available, and no personal information was explicitly extracted, stored, or used from the authors.

\subsubsection{Social Impact and Ethical Considerations}
\paragraph{Benefits}
The benefits of our dataset are two-fold: (1) \textbf{Evaluation Benchmark}: This dataset serves as a valuable evaluation benchmark for assessing the understanding of large multimodal models (LVLMs) regarding scientific articles and figures. (2) \textbf{Training Resources}: It can be used as a training resource to enhance LVLMs' understanding of scientific articles and figures, improving their performance in various scientific and research-related tasks.

\paragraph{Risks and Ethical Considerations}
However, there are potential risks and ethical considerations to address:
(1) \textbf{Misuse in Academic Integrity}: The advancement of AI research assistants facilitated by this dataset could potentially lead to misuse, such as academic fraud, fabrication, or improper assistance in academic work. We strongly encourage users to exercise caution and responsibility when using AI assistants, ensuring they are employed ethically and correctly.
(2) \textbf{Data Misinterpretation and Hallucination}: There is a risk of misinterpreting the dataset's content, leading to inaccurate conclusions or misuse of scientific information. Users should critically assess and validate the AI-generated outputs against established scientific knowledge and principles.

\subsubsection{Limitations}
Our dataset \dataset provides a comprehensive multimodal dataset across 72 scientific disciplines and serves as both a benchmark and a training resource. However, there are some limitations in our current exploration. (1) Due to limited resources, we were unable to evaluate a wide range of large-scale open-source LVLMs. (2) Our benchmark primarily assesses models' understanding of scientific figures using the figures and captions. The dataset still provide other valuable resources that could be used to create additional tasks, such as single- and multimodal questions aimed at evaluating models' scientific knowledge. We plan to explore these opportunities in future work. Despite these limitations, we believe \dataset will be a valuable resource for the research community. All data will be made publicly available.

\subsubsection{Author Statement}
The authors declare full responsibility for any rights violations, including but not limited to intellectual property rights and privacy rights, that may arise from the publication and use of this dataset. We confirm that all data provided is licensed under appropriate licenses, ensuring legal compliance and transparency.

\subsubsection{Hosting, Licensing, and Maintenance Plan}
The dataset will be hosted on GitHub, offering reliable and secure access. We commit to maintaining the repository with regular updates, security patches, and user support to ensure the data's integrity and usability over time. Licensing terms will be clearly communicated to users, adhering to the appropriate data licenses to promote proper usage and distribution.
The data is licensed under the CC BY 4.0 License, which permits sharing and adaptation with proper attribution. The primary codebase for our project is licensed under the Apache 2.0 License.

\subsection{Examples}
\begin{figure}[!ht]
\footnotesize
\centering
\includegraphics[width=0.75\textwidth]{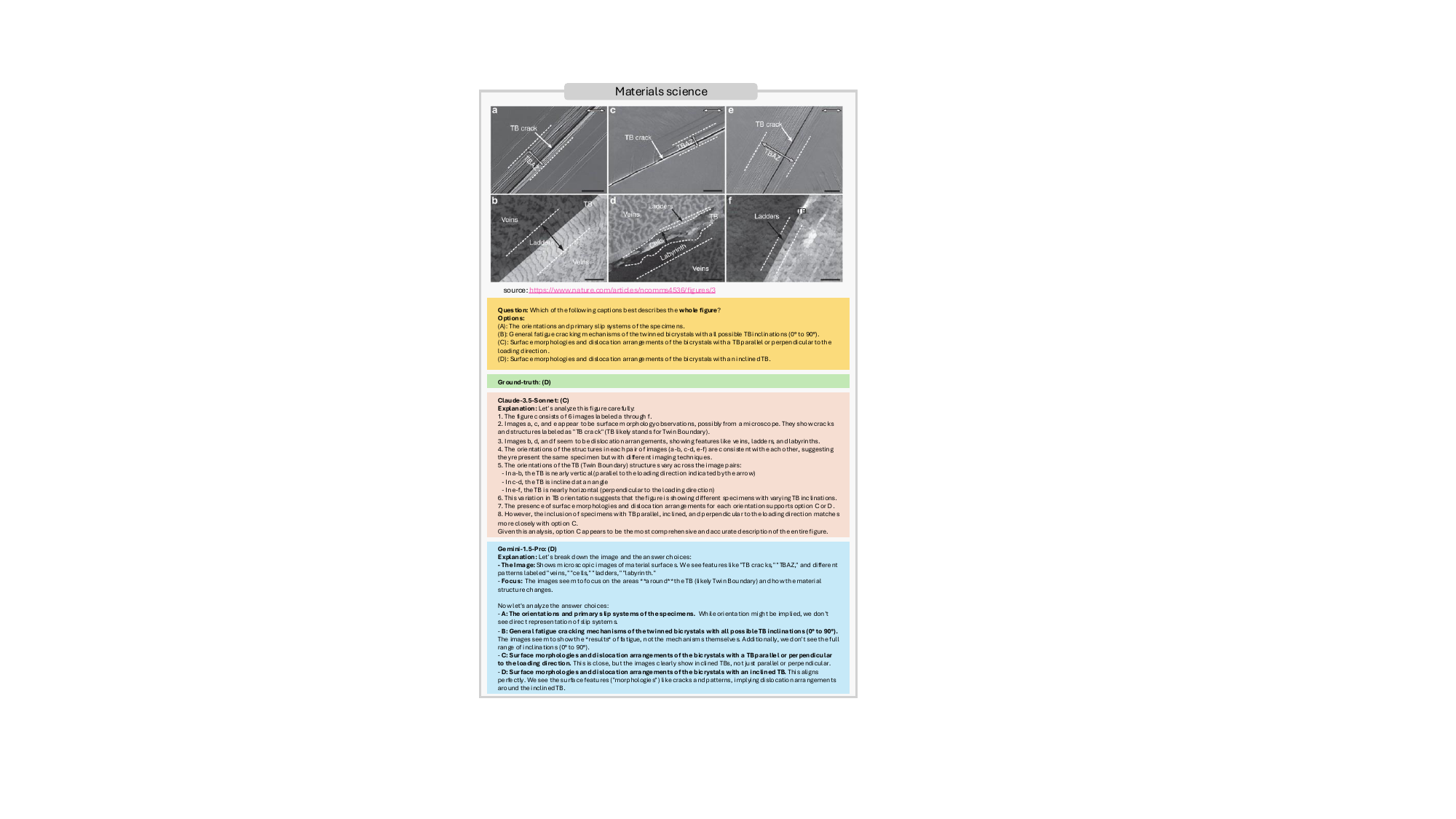}
\caption{An example of the multi-choice questions (\textbf{Fig2Cap}). The example is within the material sciences subject, sourced from \citep{setting1}. The options include the correct
main caption of the given figure and three main captions from other figures within the same article.
}
\label{fig:setting1}
\end{figure}

\begin{figure}[!ht]
\footnotesize
\centering
\includegraphics[width=0.8\textwidth]{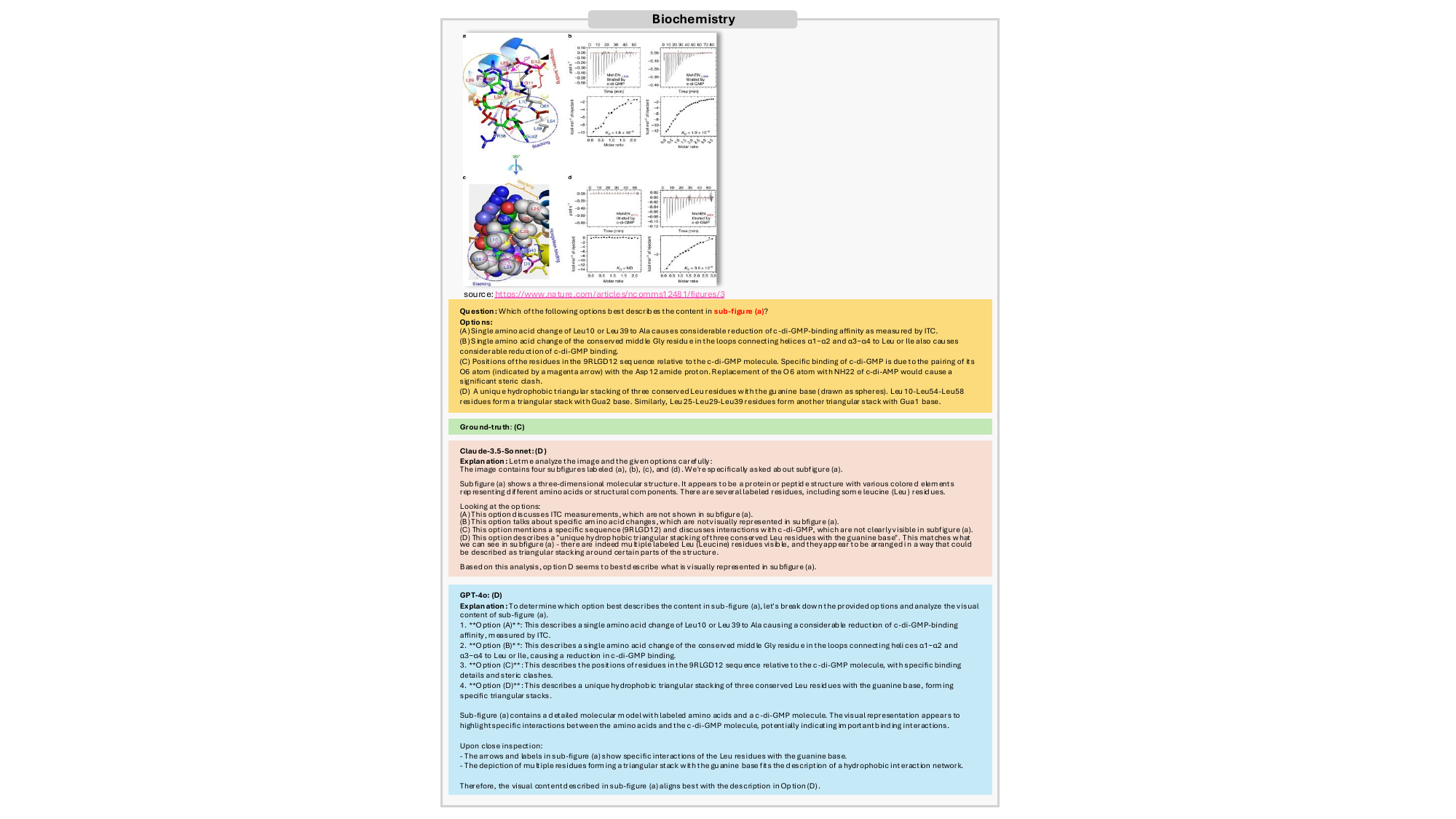}
\caption{An example of the multi-choice questions (\textbf{SubFig2Cap}). The example is within the biochemistry subject, sourced from \citep{setting3}. 
}
\label{fig:setting3}
\end{figure}

\begin{figure}[!ht]
\footnotesize
\centering
\includegraphics[width=0.75\textwidth]{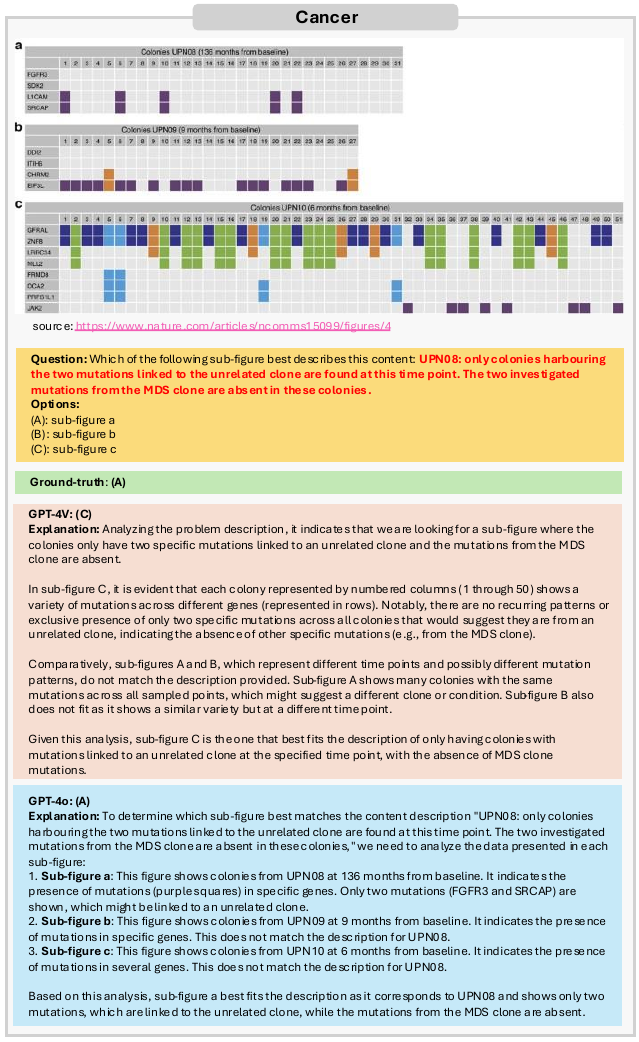}
\caption{An example of the multi-choice questions (\textbf{SubCap2Fig}). The example is within the cancer subject, sourced from \citep{setting4}. 
}
\label{fig:setting4}
\end{figure}

We present several figures as our case study to illustrate multiple-choice questions under three setting in Figure~\ref{fig:setting1}, \ref{fig:setting3}, \ref{fig:setting4}, respectively.

\end{document}